\documentclass[]{fairmeta}

\title{Intuitive physics understanding emerges from self-supervised pretraining on natural videos}

\author[1,2]{Quentin Garrido}
\author[1]{Nicolas Ballas}
\author[1]{Mahmoud Assran}
\author[1]{Adrien Bardes}
\author[1]{Laurent Najman}
\author[1]{Michael Rabbat}
\author[1,3]{Emmanuel Dupoux}
\author[1]{Yann LeCun}

\affiliation[1]{FAIR at Meta}
\affiliation[2]{Univ Gustave Eiffel}
\affiliation[3]{EHESS}

\abstract{
We investigate the emergence of intuitive physics understanding in general-purpose deep neural network models trained to predict masked regions in natural videos.
Leveraging the violation-of-expectation framework, we find that video prediction models trained to predict outcomes in a learned representation space demonstrate an understanding of various intuitive physics properties, such as object permanence and shape consistency.
In contrast, video prediction in pixel space and multimodal large language models, which reason through text, achieve performance closer to chance.
Our comparisons of these architectures reveal that jointly learning an abstract representation space while predicting missing parts of sensory input, akin to predictive coding, is sufficient to acquire an understanding of intuitive physics, and that even models trained on one week of unique video achieve above chance performance.
This challenges the idea that core knowledge --- a set of innate systems to help understand the world --- needs to be hardwired to develop an understanding of intuitive physics.
}

\date{\today}
\correspondence{Quentin Garrido at \email{garridoq@meta.com}}

\metadata[Code and data]{\url{https://github.com/facebookresearch/jepa-intuitive-physics}}

\begin{document}

\maketitle

\begin{figure}[!t]
    \centering
    \includegraphics[width=\linewidth]{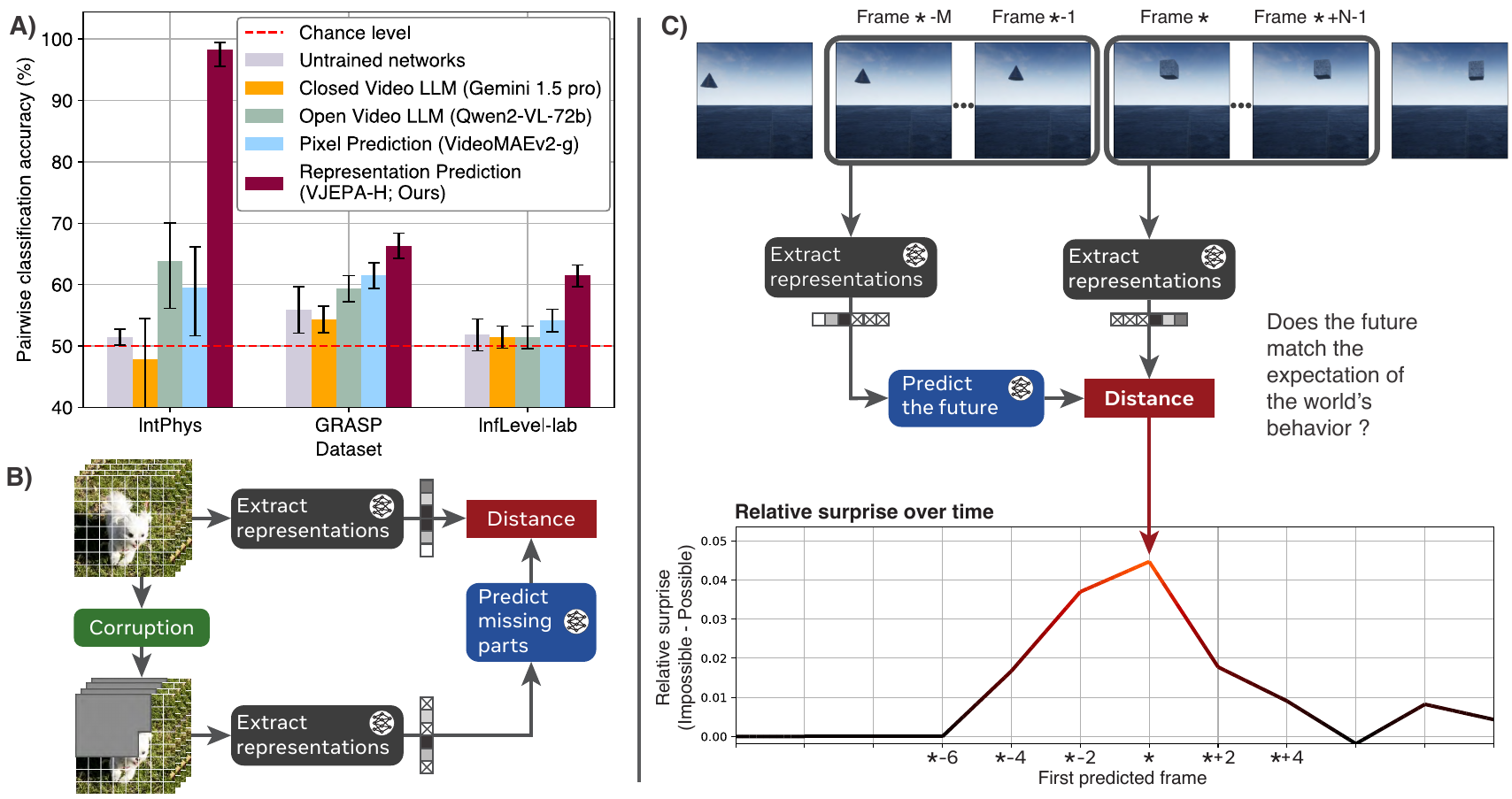}
    \caption{\textbf{Video prediction in representation space (V-JEPA) achieves an understanding of intuitive physics.} \textbf{(A)} Video models are evaluated on three intuitive physics datasets using the Violation of Expectation paradigm (IntPhys, GRASP, and InfLevel). V-JEPA is significantly more `surprised' by implausible videos. Random initializations of V-JEPA (untrained networks) show near-chance performance, and state-of-the-art video models based on text or pixel prediction are much closer to chance. Confidence intervals at 95\% are obtained via bootstrapping, except for untrained networks ($n=20$) which use a normal distribution assumption. \textbf{(B)} V-JEPA is trained to 'inpaint' natural videos in a learned representation space. Starting from a video and a corrupted version, representations are first extracted. The goal is then to predict the representation of the original video from the representation of the corrupted ones. \textbf{(C)} From a trained V-JEPA, we compute a surprise metric by predicting representations of N future frames based on M past ones and comparing the predictions to the representations of observed events. The surprise metric is then used to decide which of the two videos contains a physical violation.}
    \label{fig:figure-1}
\end{figure}

An intuitive understanding of physics is fundamental to human cognition: we expect objects to behave predictably, i.e., not to appear or disappear abruptly, move through obstacles, or change shape or color arbitrarily.
This basic grasp of the physical world has been documented not only in human infants~\citep{piaget1954construction,baillargeon_permanence_1991,baillargeon_support_1992,baillargeon_support_1990,spelke_spatiotemporal_1995}, but also in primates \citep{cacchione2004recognizing,mendes2007raising}, marine mammals \citep{singer2015object,herman2010laboratory}, corvids \citep{bird2009rooks,taylor2012new}, and chicks \citep{vallortigara2012core,wood2013newborn}. This has been taken as evidence for the \textit{core knowledge} (or core systems) hypothesis, according to which humans are equipped with a set of innate or early developing evolutionary, ancient computational systems specialized to represent and reason about basic properties of the world: objects, space, numbers, geometry, agents, etc.~\citep{baillargeon_innate_2008,spelke_core_knowledge_2007,spelke_core_2000,carey2000origin}.
In the pursuit of building machines with advanced human-level intelligence, rapid progress has produced  \emph{artificial intelligence} (AI) systems that often surpass human performance on high-level cognitive tasks like language, coding or mathematics~\citep{openai_gpt4_2024}, but paradoxically struggle in common sense physical understanding~\citep{riochet_intphys_2022,Weihs_InfLevel_2022,jassim_grasp_2024,bisk2020piqa,benchekroun2023worldsense,bansal2024videophy,bear2021physion}, 
illustrating Moravec’s paradox \citep{moravec1988mind}, namely, that tasks trivial for biological organisms can be remarkably difficult for artificial systems, and vice versa.

Previous work developing AI models with the aim of improving intuitive physics understanding can be sorted into two classes: \textit{structured models} and \textit{pixel-based generative models}. Structured models leverage hand-coded abstract representations of objects and their relationships in an Euclidean 3D space 
\citep{battaglia_simulation_2013,watters_visual_interaction_networks_2017}, yielding a powerful mental ``game engine" able to capture human's physical intuitions \citep{ullman2017mind}. This class of models can be seen as a possible computational implementation of the \textit{core knowledge} hypothesis \citep{spelke_core_knowledge_2007,spelke_core_2000}.\footnote{Weaker versions of structured models use object masks and depth cues instead of a full 3D reconstruction, e.g., \citep{riochet2020occlusion}.} Pixel-based generative models take a radically opposite view and deny the need for any hard-coded abstraction. Instead, they propose a general purpose learning mechanism consisting of reconstructing future sensory inputs (e.g., images) based on past ones~\citep{lerer2016learning,goyal2017something,finn2016unsupervised}. 

Here, we explore a third class of models that occupies a middle ground between these opposing views, integrating features from both: \emph{Joint Embedding Predictive Architectures} (JEPAs)~\citep{ lecun2022jepa,bardes_vjepa_2024}. As structured models, JEPAs posit that prediction of future world states should be done in the model's learned abstract, internal representation, and not in terms of low-level, pixel-based prediction or generation.. However, unlike structured models, JEPAs leave it to the algorithm to learn its own representation rather than hand-coding it. The mechanism consisting of predicting in representation space is congruent with the \textit{predictive coding} hypothesis of cognitive neuroscience~\citep{hohwy2013predictivemind,rao1999predictivecoding,clark2013whatever_next}. Here we study a video version of this architecture, V-JEPA~\citep{bardes_vjepa_2024}, which learns to represent video frames by reconstructing masked portions of the video in representation space.
We rely on the \emph{violation-of-expectation} framework to probe for intuitive physics understanding without requiring any task-specific training or adaptation~\citep{smith_adept_2019,riochet_intphys_2022,piloto_intuitive_2022,riochet2020occlusion}.
By prompting the model to imagine the (representation of the) future of a video and comparing its predictions with the actual observed future of the video, we obtain a quantitative measure of surprise that can be used to detect violations of intuitive physics concepts.

We find that V-JEPA accurately and consistently distinguishes between videos that follow the laws of physics and those that violate them.
Specifically, when tasked with classifying the physical plausibility of video pairs, where one video is plausible and the other is not, a V-JEPA model trained on natural videos achieves 98\% zero-shot accuracy on the IntPhys benchmark~\citep{riochet_intphys_2022} and 62\% zero-shot accuracy on the InfLevel benchmark~\citep{Weihs_InfLevel_2022}.\footnote{Here ``zero-shot'' refers both to the fact that the V-JEPA models were not trained specifically for the task of distinguishing between physically plausible and implausible videos, and that the model was not trained on data from any of the benchmarks.}
Surprisingly, we find that multimodal large-language models~\citep{wang_2024_qwen,reid2024gemini} and comparable video prediction methods making predictions in pixel-space~\citep{wang_videomaev2_2023} perform around chance.

To better understand which design choices lead to the emergence of intuitive physics understanding in V-JEPA, we ablate the effect of the training data, the pretraining prediction objective (what to predict from what), and the model size.
While we observe that varying each of these components influences performance, all V-JEPA models achieve performance significantly above chance, including a small 115 million parameter model, or a model trained on only one week of unique video, thereby suggesting that video prediction in a learned representation space is a robust objective for acquiring intuitive physics understanding.

\section*{Measuring intutive physics understanding}

\textbf{Violation of Expectation.}
The violation-of-expectation paradigm has its roots in developmental psychology~\citep{margoni_voe_2024,baillargeon1985object}.
Subjects, typically infants, are presented with two similar visual scenes, one of which contains a physical impossibility.
A `surprise' reaction to each scene is then obtained through various physiological measures, such as relative gaze time~\citep{spelke1985voe_looking}, and is used to determine whether a concept violation has occurred in the subject~\citep{baillargeon_permanence_1991,spelke1985voe_looking, margoni_voe_2024}.\footnote{Non-conceptual interpretations of gaze-times, e.g., based on low-level processes such as perceptual preferences, are typically mitigated to some degree in these experiments by conducting a series of habituation trials prior to the violation-of-expectation trials.}
This paradigm has been extended to evaluate the physical understanding of AI systems~\citep{riochet_intphys_2022,smith_adept_2019,riochet2020occlusion}, where, similarly to infant trials, pairs of scenes are presented to a model with all aspects (properties of objects, number of objects, occluders, etc.) kept identical across the two scenes, apart from a single aspect or event that violates a specific intuitive physics concept.
For example, a ball may roll behind an occluder but never reappear in one of the paired videos, thereby testing for the concept of object permanence.
A higher surprise response attributed by the model to the impossible scenario reflects a correct understanding of the violated concept.

\textbf{Video Prediction for intuitive physics understanding.}
The V-JEPA architecture~\citep{lecun2022jepa} has been primarily developed to improve the capacity of a model to adapt to high-level downstream tasks, such as activity recognition~\citep{kay2017kinetics} and action classification~\citep{goyal2017ssv2}, directly from the input without hard-wiring a cascade of intermediate representations like object contours or pose estimation \citep{bardes_vjepa_2024}. Here, we test the hypothesis that the reason this architecture is successful at high-level tasks is that it has learned a representation that implicitly captures the structure and dynamics of objects in the world without the need to represent them directly.

As illustrated in Figure~\ref{fig:figure-1}.B, V-JEPA is instantiated with an encoder (a neural network) that extracts representations from a video, and a predictor (also a neural network) that predicts the representation of an artificially masked part of the video, such as a randomly masked spatiotemporal block, random pixels, or future frames.
This joint training of the encoder and predictor enables the encoder to learn abstract representations that encode predictable information and discard low-level (typically less semantic) features.
Refer to Section~\ref{sec:method} in the supplementary material for more details on architecture and training.

After self-supervised training, we can use the encoder and predictor networks, without any additional adaptation, to probe the model's understanding of the world.
Specifically, iterating through a stream of video, the model encodes the observed pixels and subsequently predicts the representation of the following frames in the video, as illustrated in Figure~\ref{fig:figure-1}.C.
By recording the \emph{prediction error} --- the distance between the predicted video representations and the actual encoded video representations --- at each time-step, we obtain a temporally aligned quantitative measure of the model's surprise throughout the video.
Varying how many past video frames (context) a model can use to predict the future allows us to control for memory, while varying the frame rate of the video allows us to control for the fineness of motions.
Refer to Section~\ref{sec:surprise} in the supplementary material for more details.

\section*{Representation prediction learns to detect violations of intuitive physics}

We evaluate intuitive physics understanding on three datasets:~the dev set of IntPhys~\citep{riochet_intphys_2022}, GRASP~\citep{jassim_grasp_2024} and InfLevel-lab~\citep{Weihs_InfLevel_2022}.
This mix of benchmarks provides diversity in the visual quality (synthetic/photorealistic), in the diversity of scenes considered, as well as in the intuitive physics properties that are probed.
Specifically, the combination of these datasets allows us to probe the understanding of object permanence~\citep{baillargeon_permanence_1991}, continuity~\citep{spelke_origins_1992}, shape and color constancy~\citep{wilcox1999constancy,wilcox2004constancy}, gravity~\citep{kim1992gravity}, support~\citep{baillargeon_support_1990,baillargeon_support_1992}, solidity~\citep{spelke_origins_1992}, inertia~\citep{spelke_origins_1992}, and collision~\citep{baillargeon1995collision}. See Section~\ref{sec:properties} in the supplementary material for exact definitions.

We compare V-JEPA to other video models to investigate how important to intuitive physics understanding is the video prediction objective, as well as the representation space where prediction is performed.
We consider two other classes of models: video prediction models that predict directly in pixel space, and Multimodal Large Language Models (MLLMs).
The former set of pre-training methods have a similar prediction objective as V-JEPA, but often learn representation spaces with poor semanticity~\citep{wang_videomaev2_2023,bardes_vjepa_2024}; they are useful once fine-tuned for a specific task. As a representative method, we evaluate VideoMAEv2~\citep{wang_videomaev2_2023}. While different prediction objectives and pretraining data are used, this allows a comparison to V-JEPA in terms of prediction space. Given its predictive nature, VideoMAEv2 can be evaluated in the same way as V-JEPA, by predicting the future and measuring surprise via prediction error.

The latter class of models, MLLMs, are trained to predict text and are only interleaved with video a posteriori, making them devoid of a video prediction objective. As exemplar methods, we study Qwen2-VL-7B~\citep{wang_2024_qwen}, a state-of-the-art, open-weights, video-language model, and Gemini 1.5 pro~\citep{reid2024gemini}, a closed commercial model. These models are both significantly larger than V-JEPA in terms of parameter count and the amount of data they were trained on, and they learn primarily from text data.
Multimodal LLMs take videos and potentially a text prompt as input and learn to generate a corresponding textual output.
Due to their text-only output, those models cannot use the same evaluation protocol based on a quantitative measure of surprise. Instead, we give the model a pair of videos, asking which one of the two is impossible. Section~\ref{sec:surprise} in the supplementary material describes the detailed protocol. 

For every method considered, we evaluate the flagship models proposed in the original works. We further compare all models with untrained neural networks, testing the learnability of intuitive physics understanding. For each property and model, the context size is chosen as the one maximizing performance, allowing the models to adapt to the different evaluation setups. This process is done for all methods, and leads to results illustrating the best performance achievable by the model. We expand on this choice in section~\ref{sec:context} in the supplementary material.

We summarize the performance of methods across datasets on pairwise classification (i.e., detecting the impossible video in a pair) in Figure~\ref{fig:figure-1}.A. Refer to Section~\ref{sec:complete_all_methods} in the supplementary material for detailed results, and Section~\ref{sec:hyperparams} for detailed parameters used.

We find that V-JEPA is the only method that achieves significantly higher performance than untrained networks across all datasets, achieving average accuracies of 98\% (95\% CI [95\%,99\%]), 66\% (95\% CI [64\%,68\%]) , 62\% (95\% CI [60\%,63\%]) respectively on IntPhys, GRASP, and InfLevel-lab. These results show that prediction in a learned representation space is sufficient to develop an understanding of intuitive physics. This is done without any predefined abstractions, and without knowledge of the benchmarks during pretraining or development of the method. 

By comparison, we find that VideoMAEv2, Qwen2-VL-7B, and Gemini 1.5 pro achieve performance that is only marginally above that of randomly-initialized models. The low performance of pixel prediction and multimodal LLMs corroborates previous findings~\citep{riochet_intphys_2022,jassim_grasp_2024}. These comparisons further highlight the benefit of V-JEPA over the existing VideoMAEv2, Gemini 1.5 pro, and Qwen2-VL-72B models.
These results, however, do not mean that LLMs or pixel prediction models cannot achieve intuitive physics understanding, but merely that this seemingly simple task remains difficult even for frontier models~\citep{jassim_grasp_2024,kang2024farvideogenerationworld,bansal2024videophy}.

\begin{figure}[!t]
    \centering
    \includegraphics[width=1\linewidth]{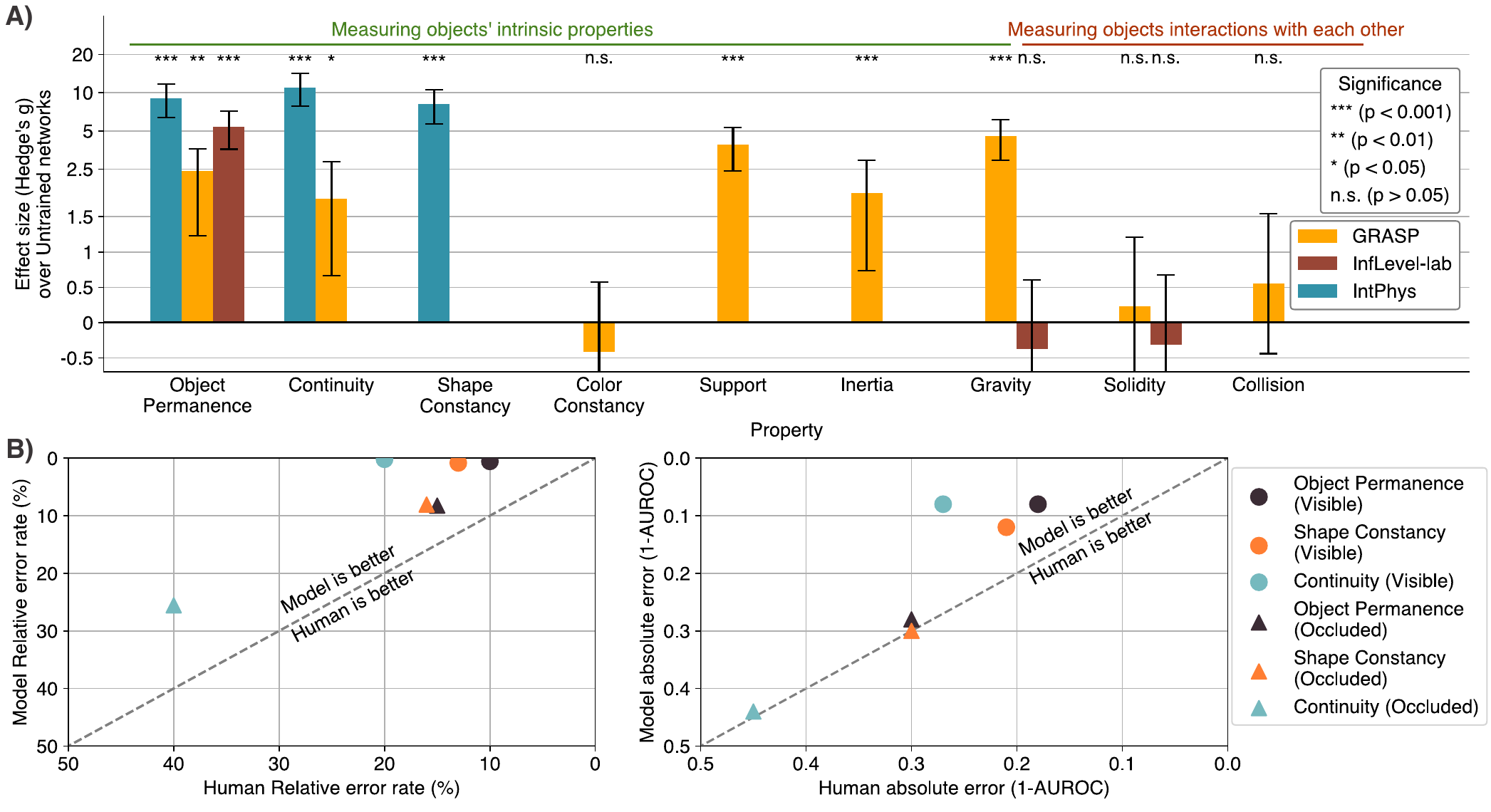}
    \caption{\textbf{V-JEPA accuracy increase relative to randomly-initialized models and humans across different physical properties and benchmarks.} \textbf{(A)} Because some benchmarks contain low-level biases, we test the model performance against a set of randomly initialized networks ($n=20$). V-JEPA models ($n=5$) have higher relative classification accuracy on intuitive physics benchmarks for most, but not all concepts. \textbf{(B)} V-JEPA relative (left) and absolute (right) accuracy on the IntPhys test set across different conditions compared to naive human performance, showing a high correlation between human and machine errors. The V-JEPA score uses the maximum surprise from each video, which generalizes better for single-video classification.  Human data are taken from~\citep{riochet_intphys_2022}.}
    \label{fig:all_results}
\end{figure}

\section*{Per property analysis of V-JEPA}

We now take a closer look at the per-property performance of V-JEPA on the previously used datasets in order to obtain a more precise understanding of its intuitive physics understanding.
Here, the V-JEPA encoder and predictor are based on the Vision Transformer-Large (ViT-L, instead of ViT-H for the flagship model)~\citep{dosovitskiy2021an,bardes_vjepa_2024} architecture and are trained on the HowTo100M dataset~\citep{miech2019howto100m}. We perform a two-sample one-tailed Welch's t-test to assess whether V-JEPA (n=5) provides increased performance over randomly-initialized, untrained models (n=20). 
The results are summarized in Figure~\ref{fig:all_results}.

On IntPhys, we find V-JEPA to significantly outperform untrained networks on multiple intuitive physics properties: Object Permanence: M=85.7, SD=7.6 vs.~M=51.4, SD=1.0 (t(4.0) = -8.9, $p$ = $4.19\times10^{-4}$), with an effect size $g$ = 9.0 (95\% CI [6.3,11.7]); Continuity: M=86.3, SD=6.2 vs.~M=51.2, SD=1.2 (t(4.1) = -11.3, $p$ = $1.61\times10^{-4}$), with an effect size $g$ = 11.0 (95\% CI [7.8,14.2]); Shape Constancy: M=83.7, SD=7.8 vs.~M=51.7, SD=1.2 (t(4.0) = -8.1, $p$ = $5.96\times10^{-4}$), with an effect size $g$ = 8.1 (95\% CI [5.7,10.6]).
On GRASP, we find significantly higher accuracies for V-JEPA on: Object Permanence: M=70.7, SD=7.8 vs.~M=54.1, SD=5.9 (t(5.0) = -4.0, $p$ = $5.10\times10^{-3}$), with an effect size $g$ = 2.4 (95\% CI [1.2,3.6]); Continuity: M=65.0, SD=6.1 vs.~M=55.0, SD=5.0 (t(5.2) = -3.0, $p$ = $1.36\times10^{-2}$), with an effect size $g$ = 1.8 (95\% CI [0.7,2.9]); Support: M=98.1, SD=3.0 vs.~M=58.4, SD=10.5 (t(21.4) = -14.0, $p$ = $1.48\times10^{-12}$), with an effect size $g$ = 3.9 (95\% CI [2.4,5.3]); Gravity: M=74.9, SD=2.4 vs.~M=55.3, SD=4.3 (t(10.3) = -12.6, $p$ = $6.83\times10^{-8}$), with an effect size $g$ = 4.5 (95\% CI [2.9,6.1]); Inertia: M=62.0, SD=2.4 vs.~M=54.3, SD=4.2 (t(10.1) = -5.1, $p$ = $2.36\times10^{-4}$), with an effect size $g$ = 1.8 (95\% CI [0.7,2.9]).
However, we do not find a significant gain on: Color Constancy, Solidity, or Collision ($p > 0.05$).
On InfLevel, we find significantly higher accuracies for V-JEPA on: Object Permanence: M=72.1, SD=2.9 vs.~M=52.5, SD=3.5 (t(6.8) = -11.9, $p$ = $4.46\times10^{-6}$), with an effect size $g$ = 5.4 (95\% CI [3.6,7.1]). However, we do not find a significant gain on: Gravity or Solidity ($p > 0.05$).

V-JEPA excels at properties related to the scene's content (e.g., object permanence), but struggles with categories that require knowledge of a contextualizing event (gravity and solidity in InfLevel-lab) or the modeling of precise object interactions such as collisions. We hypothesize that these limitations come mainly from the model's framerate constraints.
Nevertheless, V-JEPA demonstrates an understanding of intuitive physics while learning the required abstractions from the raw perceptual signal and without relying on strong prior information. In contrast to previous work~\citep{smith_adept_2019,riochet_intphys_2022}, this suggests that core knowledge is not necessary for deep learning systems to understand intuitive physics concepts.

We further compare V-JEPA to human performance using the private test set from IntPhys~\citep{riochet_intphys_2022}. The human data is taken from~\citep{riochet_intphys_2022,riochet2020occlusion}, where it was obtained through Amazon Mechanical Turk. For this experiment, we focus on the flagship V-JEPA architecture, using a ViT-Huge~\citep{dosovitskiy2021an,bardes_vjepa_2024} with pretraining on VideoMix2M \citep{bardes_vjepa_2024}. 
We find that V-JEPA achieves equal or higher performance for all intuitive physics properties, as illustrated in Figure~\ref{fig:all_results}.B. 
We find that using the maximum surprise in a video, rather than the average, leads to better performance on single videos. We discuss further this distinction in Section~\ref{sec:surprise} in the supplementary material.
In general,  we observe lower performance in both V-JEPA and humans for videos where the physics-breaking event happens behind an occluder. Additionally, performance is well-correlated between humans and V-JEPA for the occluded settings.

\section*{Keys to intuitive physics understanding}

\begin{figure}
    \centering
    \includegraphics[width=1\linewidth]{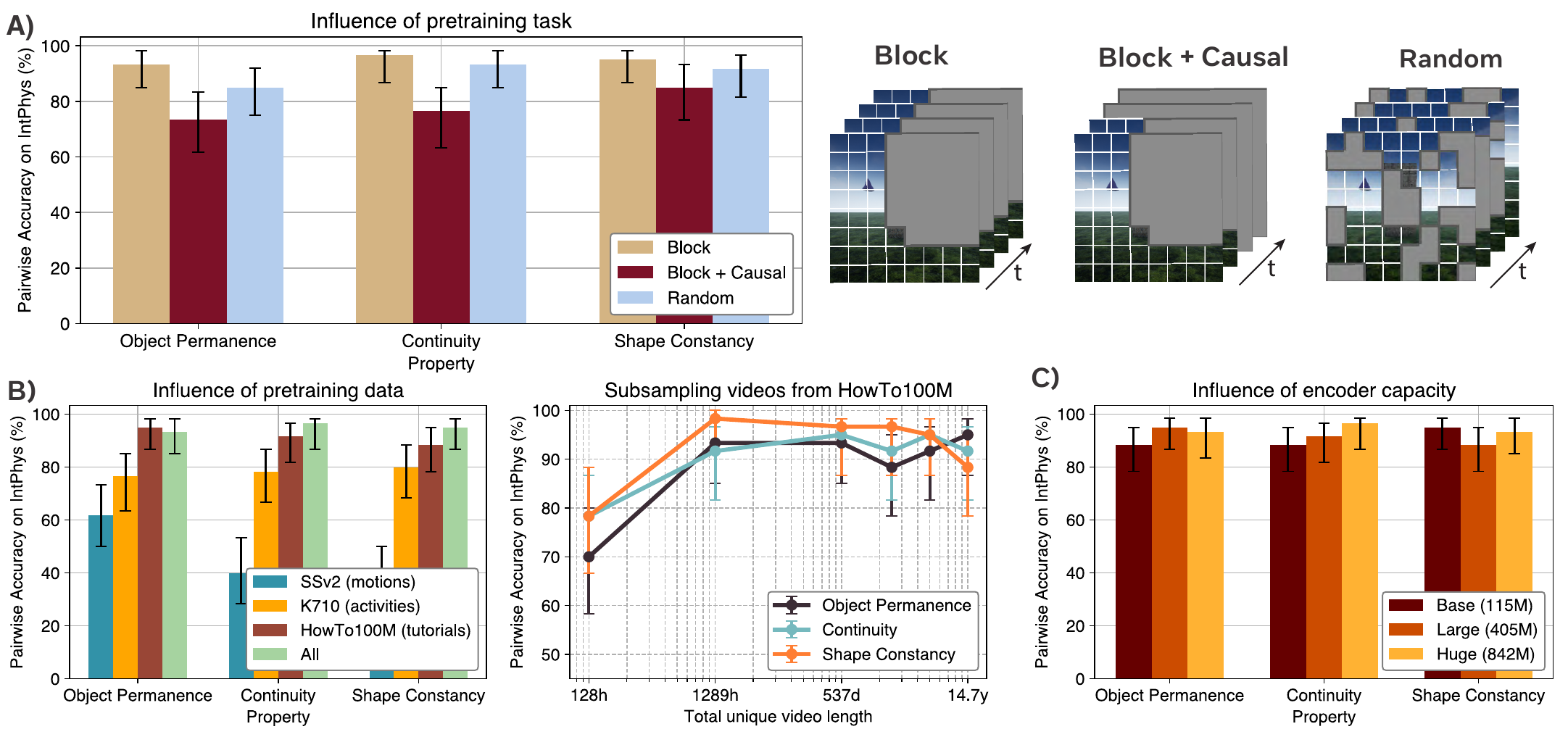}
    \caption{\textbf{Influence of type of mask, type and amount of training data, and model size on V-JEPA IntPhys scores.} \textbf{(A)} When pretrained on VM2M, V-JEPA exhibits an understanding of intuitive physics with every masking strategy. \textbf{(B)} Of the three training datasets, two give high accuracies when trained separately (K710 and Howto100M). High scores are found with only 1289 hours of Howto100M (the largest dataset), and even 128h gives better than chance performance. \textbf{(C)} While larger encoders improve performance, we find that the performance remains non-trivial across sizes when pretraining on HowTo100M. Confidence intervals obtained via bootstrapping.}
    \label{fig:ablations}
\end{figure}

We now ablate V-JEPA design choices to better understand the conditions for intuitive physics understanding to emerge. We focus on three components that play a crucial role in the model's capabilities.
First, we examine the impact of the training data. The choice of data defines the learning environment of the model, with different video sources providing variations in semantic diversity, movement patterns, and quantity. Second, we consider the effect of the model size. While conventional wisdom states that larger models perform better, we also ponder the minimum size required to achieve non-trivial performance.
Third, we study the influence of the pretraining prediction task.
Does selecting what to predict from what observed context (pretraining masking strategy) affect the model's understanding of intuitive physics?

\textbf{Importance of the pretraining task.} Recall that V-JEPA models are trained to predict representations of randomly masked portions of a video, but always perform causal prediction at inference time, where the context includes frames up to some time $t$ and the model should predict representations of frames at times greater than $t$. 
Although we compute V-JEPA's surprise using causal prediction and have observed above that this is effective for intuitive physics understanding, V-JEPA is never trained using a causal prediction task.
Rather, the pre-training task is referred to as \emph{Block Masking}~\citep{bardes_vjepa_2024}, where a large spatial block is masked for the full duration of the video. V-JEPA's performance on action and activity recognition tasks has previously been observed to vary drastically depending on the exact strategy used~\citep{bardes_vjepa_2024}.

To understand the extent to which V-JEPA intuitive physics understanding emerges specifically from the \textit{Block Masking} training task, we study the effect of changing this training task, and consider two possible alternatives. \textit{Causal Block Masking} is similar to \textit{Block Masking}, but also fully masks the last 25\% of the video, thereby incorporating future prediction into the training procedure, and \textit{Random Masking} which masks random pixels in the video.
Contrary to classical video tasks~\citep{bardes_vjepa_2024}, we find that the prediction task is not as important for intuitive physics understanding (see figure \ref{fig:ablations}.B).
Whereas \textit{Random Masking} leads to a drop of 20 points on average on video classification tasks~\citep{bardes_vjepa_2024}, the drop on IntPhys is only around 5 points on average.
Interestingly, \textit{Causal Block Masking} seems to perform worse than its non-causal counterpart, despite being more closely aligned to the model's prediction setup at test time.
The effective performance of \textit{Random Masking}, perhaps the simplest strategy, suggests that the understanding of intuitive physics does not require a tailored objective, but that predicting in an abstract representation space is the key aspect.

\textbf{Importance of pre-training data.}
Data is a key ingredient of deep learning models and video models are no exception~\citep{bardes_vjepa_2024}.
Video datasets can be described along several axes, such as the number of distinct videos, the (average) duration of videos, whether videos are captured from egocentric or exocentric views, whether that camera is static or moving, and so on.
We thus investigate in more detail the influence of pretraining data on intuitive physics performance.
V-JEPA has previously been trained on a mixture of three popular video datasets, referred to as VideoMix2M \citep{bardes_vjepa_2024}: Kinetics 710 (K710~\citep{kay2017kinetics}), Something-Something-v2 (SSv2~\citep{goyal2017ssv2}) and HowTo100M (HowTo~\citep{miech2019howto100m}).
Each of these datasets focuses on a different slice of the distribution of natural videos, namely, activities in K710 (e.g., playing basketball), fine-grained motion in SSv2 (e.g., throwing something), and tutorials in HowTo100M (e.g., cooking). 
To study the influence of training data on learning intuitive physics, we re-train V-JEPA-L models separately using only one of the three component datasets.

Unsurprisingly, we find a strong impact of data sources on performance. Training only with videos based on motion understanding (SSv2) leads to almost chance-level performance. While more action-focused data (K710) leads to an above-chance understanding of intuitive physics, we find that tutorial videos (HowTo) yield the best performance among individual component datasets. 
However, HowTo is also larger than SSv2 and K710 (15 years vs.~3 months combined).
We thus further examine the evolution of performance with smaller datasets coming from the same distribution by subsampling HowTo100M.
We hold the compute budget fixed across these experiments such that model training always processes the equivalent of 30 years of video (by revisiting videos from the training dataset multiple times) even when only using 0.1\% of HowTo100M, which represents only 128 hours of unique video in total.
We find in Figure~\ref{fig:ablations}.C that the size of the dataset does not meaningfully impact performance, and that the model can adequately distinguish violations of intuitive physics concepts even with 128h of unique videos, maintaining a pairwise accuracy of over 70\% on all considered properties.

\textbf{Importance of the encoder size.}
Common wisdom in the deep learning literature is that larger models perform better~\citep{kaplan2020scaling}. Here, we are also interested in the minimal size at which we observe evidence of non-trivial intuitive physics understanding. We thus investigate what happens in both directions of the scaling, using smaller and larger encoders.
In Figure~\ref{fig:ablations}.C, we find that larger models tend to perform better. However, a 115 million parameters model still achieves an accuracy of over 85\%, demonstrating a robust understanding of intuitive physics.

\section*{Discussion}

In this work, we studied the emergence of intuitive physics understanding in state-of-the-art deep learning models.
By pretraining on natural videos with a simple prediction task in a learned representation space, V-JEPA exhibits an understanding of intuitive physics on both synthetic and real videos without any task-specific adaptation.
Our results show that intuitive physics understanding can be acquired using a general learning principle, and thus does not require hardwired \textit{core knowledge}. Although we find that the size of the model, the choice of pretraining data, and the exact pretraining task influence this understanding, its emergence can be attributed to the general framework of representation space prediction rather than a precise design choice of V-JEPA. When studying other methods such as multimodal LLMs and pixel prediction methods, we find that current models perform around chance level. Higher-capacity generative video models could potentially benefit from a certain understanding of intuitive physics~\citep{brooks2024sora} in order to produce realistic videos. Yet, current evidence points to an incomplete understanding of physics in existing video generative models~\citep{motamed2025physicsiq,bansal2024videophy}\footnote{Most state-of-the-art models~\citep{brooks2024sora} being proprietary complicates a rigorous assessment of their physics understanding due to their lack of openness.}.\\
Nonetheless, the demonstrated understanding of V-JEPA is not without limitations. 
Indeed, V-JEPA is not uniformly accurate under all conditions. Figure 2 shows that although the accuracies are high for physical violations that imply properties intrinsic to objects (except for the color property), violations implicating interactions between objects, like solidity or collision, are close to chance. This may be due to the fact that object interactions are not very frequent in the model training data, and are not learned as well as more frequent ones. Furthermore, current JEPA models have limited memory, and consequently process very short video clips at a time (typically 3--4 seconds). V-JEPA also lacks the ability to condition its predictions on additional context, such as an action taking place, and thus predicts the future only as an observer. Although this lends itself well to the tested properties, more complex interactions are out of reach at the moment.
Indeed, it could be that interactions between objects require higher-order representations, and that a more powerful hierarchical version of JEPA is needed to capture these interactions. Finally, it is also possible that an agent has to be able to interact with objects themselves in order to learn about interactions, suggesting the need to add action channels to the learning system. \\
From a data standpoint, it would also be interesting to study models trained on videos that mimic what infants see~\citep{sullivan2021saycam, long_babyview_2024}, and whether an understanding of intuitive physics also emerges in models trained on such data.\\
Nonetheless, through the results reported here, we believe that the latent prediction framework is a path forward toward building neural networks that understand the physical world.

\newpage
\bibliographystyle{assets/plainnat}
\bibliography{paper}

\begin{thebibliography}{64}
\providecommand{\natexlab}[1]{#1}
\providecommand{\url}[1]{\texttt{#1}}
\expandafter\ifx\csname urlstyle\endcsname\relax
  \providecommand{\doi}[1]{doi: #1}\else
  \providecommand{\doi}{doi: \begingroup \urlstyle{rm}\Url}\fi

\bibitem[Baillargeon(1995)]{baillargeon1995collision}
Renee Baillargeon.
\newblock Physical reasoning in infancy.
\newblock \emph{The cognitive neurosciences}, pages 181--204, 1995.

\bibitem[Baillargeon and DeVos(1991)]{baillargeon_permanence_1991}
Renee Baillargeon and Julie DeVos.
\newblock Object {Permanence} in {Young} {Infants}: {Further} {Evidence}.
\newblock \emph{Child Development}, 62\penalty0 (6):\penalty0 1227, December
  1991.
\newblock ISSN 00093920.
\newblock \doi{10.2307/1130803}.

\bibitem[Baillargeon et~al.(1985)Baillargeon, Spelke, and
  Wasserman]{baillargeon1985object}
Renee Baillargeon, Elizabeth~S Spelke, and Stanley Wasserman.
\newblock Object permanence in five-month-old infants.
\newblock \emph{Cognition}, 20\penalty0 (3):\penalty0 191--208, 1985.

\bibitem[Baillargeon(2008)]{baillargeon_innate_2008}
Renée Baillargeon.
\newblock Innate {Ideas} {Revisited}: {For} a {Principle} of {Persistence} in
  {Infants}' {Physical} {Reasoning}.
\newblock \emph{Perspectives on Psychological Science}, 3\penalty0
  (1):\penalty0 2--13, January 2008.
\newblock ISSN 1745-6916, 1745-6924.
\newblock \doi{10.1111/j.1745-6916.2008.00056.x}.

\bibitem[Baillargeon and Hanko-Summers(1990)]{baillargeon_support_1990}
Renée Baillargeon and Stephanie Hanko-Summers.
\newblock Is the top object adequately supported by the bottom object? young
  infants' understanding of support relations.
\newblock \emph{Cognitive Development}, 5\penalty0 (1):\penalty0 29--53,
  January 1990.
\newblock ISSN 08852014.
\newblock \doi{10.1016/0885-2014(90)90011-H}.

\bibitem[Baillargeon et~al.(1992)Baillargeon, Needham, and
  Devos]{baillargeon_support_1992}
Renée Baillargeon, Amy Needham, and Julie Devos.
\newblock The development of young infants' intuitions about support.
\newblock \emph{Early Development and Parenting}, 1\penalty0 (2):\penalty0
  69--78, January 1992.
\newblock ISSN 1057-3593, 1099-0917.
\newblock \doi{10.1002/edp.2430010203}.

\bibitem[Bansal et~al.(2024)Bansal, Lin, Xie, Zong, Yarom, Bitton, Jiang, Sun,
  Chang, and Grover]{bansal2024videophy}
Hritik Bansal, Zongyu Lin, Tianyi Xie, Zeshun Zong, Michal Yarom, Yonatan
  Bitton, Chenfanfu Jiang, Yizhou Sun, Kai-Wei Chang, and Aditya Grover.
\newblock Videophy: Evaluating physical commonsense for video generation.
\newblock \emph{arXiv:2406.03520}, 2024.

\bibitem[Bardes et~al.(2024)Bardes, Garrido, Ponce, Chen, Rabbat, LeCun,
  Assran, and Ballas]{bardes_vjepa_2024}
Adrien Bardes, Quentin Garrido, Jean Ponce, Xinlei Chen, Michael Rabbat, Yann
  LeCun, Mido Assran, and Nicolas Ballas.
\newblock Revisiting feature prediction for learning visual representations
  from video.
\newblock \emph{Transactions on Machine Learning Research}, 2024.
\newblock ISSN 2835-8856.
\newblock Featured Certification.

\bibitem[Battaglia et~al.(2013)Battaglia, Hamrick, and
  Tenenbaum]{battaglia_simulation_2013}
Peter~W. Battaglia, Jessica~B. Hamrick, and Joshua~B. Tenenbaum.
\newblock Simulation as an engine of physical scene understanding.
\newblock \emph{Proceedings of the National Academy of Sciences}, 110\penalty0
  (45):\penalty0 18327--18332, November 2013.
\newblock ISSN 0027-8424, 1091-6490.
\newblock \doi{10.1073/pnas.1306572110}.

\bibitem[Bear et~al.(2021)Bear, Wang, Mrowca, Binder, Tung, Pramod, Holdaway,
  Tao, Smith, Sun, et~al.]{bear2021physion}
Daniel~M Bear, Elias Wang, Damian Mrowca, Felix~J Binder, Hsiao-Yu~Fish Tung,
  RT~Pramod, Cameron Holdaway, Sirui Tao, Kevin Smith, Fan-Yun Sun, et~al.
\newblock Physion: Evaluating physical prediction from vision in humans and
  machines.
\newblock \emph{arXiv:2106.08261}, 2021.

\bibitem[Benchekroun et~al.(2023)Benchekroun, Dervishi, Ibrahim, Gaya,
  Martinet, Mialon, Scialom, Dupoux, Hupkes, and
  Vincent]{benchekroun2023worldsense}
Youssef Benchekroun, Megi Dervishi, Mark Ibrahim, Jean-Baptiste Gaya, Xavier
  Martinet, Gr{\'e}goire Mialon, Thomas Scialom, Emmanuel Dupoux, Dieuwke
  Hupkes, and Pascal Vincent.
\newblock Worldsense: A synthetic benchmark for grounded reasoning in large
  language models.
\newblock \emph{arXiv:2311.15930}, 2023.

\bibitem[Bird and Emery(2009)]{bird2009rooks}
Christopher~David Bird and Nathan~John Emery.
\newblock Rooks use stones to raise the water level to reach a floating worm.
\newblock \emph{Current Biology}, 19\penalty0 (16):\penalty0 1410--1414, 2009.

\bibitem[Bisk et~al.(2020)Bisk, Zellers, Gao, Choi, et~al.]{bisk2020piqa}
Yonatan Bisk, Rowan Zellers, Jianfeng Gao, Yejin Choi, et~al.
\newblock Piqa: Reasoning about physical commonsense in natural language.
\newblock \emph{Proceedings of the AAAI conference on artificial intelligence},
  34:\penalty0 7432--7439, 2020.

\bibitem[Brooks et~al.(2024)Brooks, Peebles, Holmes, DePue, Guo, Jing, Schnurr,
  Taylor, Luhman, Luhman, Ng, Wang, and Ramesh]{brooks2024sora}
Tim Brooks, Bill Peebles, Connor Holmes, Will DePue, Yufei Guo, Li~Jing, David
  Schnurr, Joe Taylor, Troy Luhman, Eric Luhman, Clarence Ng, Ricky Wang, and
  Aditya Ramesh.
\newblock Video generation models as world simulators, 2024.

\bibitem[Cacchione and Krist(2004)]{cacchione2004recognizing}
Trix Cacchione and Horst Krist.
\newblock Recognizing impossible object relations: intuitions about support in
  chimpanzees (pan troglodytes).
\newblock \emph{Journal of Comparative Psychology}, 118\penalty0 (2):\penalty0
  140, 2004.

\bibitem[Carey(2000)]{carey2000origin}
Susan Carey.
\newblock The origin of concepts.
\newblock \emph{Journal of Cognition and Development}, 1\penalty0 (1):\penalty0
  37--41, 2000.

\bibitem[Clark(2013)]{clark2013whatever_next}
Andy Clark.
\newblock Whatever next? predictive brains, situated agents, and the future of
  cognitive science.
\newblock \emph{Behavioral and brain sciences}, 36\penalty0 (3):\penalty0
  181--204, 2013.

\bibitem[Dosovitskiy et~al.(2021)Dosovitskiy, Beyer, Kolesnikov, Weissenborn,
  Zhai, Unterthiner, Dehghani, Minderer, Heigold, Gelly, Uszkoreit, and
  Houlsby]{dosovitskiy2021an}
Alexey Dosovitskiy, Lucas Beyer, Alexander Kolesnikov, Dirk Weissenborn,
  Xiaohua Zhai, Thomas Unterthiner, Mostafa Dehghani, Matthias Minderer, Georg
  Heigold, Sylvain Gelly, Jakob Uszkoreit, and Neil Houlsby.
\newblock An image is worth 16x16 words: Transformers for image recognition at
  scale.
\newblock \emph{International Conference on Learning Representations}, 2021.

\bibitem[Finn et~al.(2016)Finn, Goodfellow, and Levine]{finn2016unsupervised}
Chelsea Finn, Ian Goodfellow, and Sergey Levine.
\newblock Unsupervised learning for physical interaction through video
  prediction.
\newblock \emph{Advances in neural information processing systems}, 29, 2016.

\bibitem[Goyal et~al.(2017{\natexlab{a}})Goyal, Ebrahimi~Kahou, Michalski,
  Materzynska, Westphal, Kim, Haenel, Fruend, Yianilos, Mueller-Freitag,
  et~al.]{goyal2017something}
Raghav Goyal, Samira Ebrahimi~Kahou, Vincent Michalski, Joanna Materzynska,
  Susanne Westphal, Heuna Kim, Valentin Haenel, Ingo Fruend, Peter Yianilos,
  Moritz Mueller-Freitag, et~al.
\newblock The" something something" video database for learning and evaluating
  visual common sense.
\newblock \emph{Proceedings of the IEEE international conference on computer
  vision}, pages 5842--5850, 2017{\natexlab{a}}.

\bibitem[Goyal et~al.(2017{\natexlab{b}})Goyal, Ebrahimi~Kahou, Michalski,
  Materzynska, Westphal, Kim, Haenel, Fruend, Yianilos, Mueller-Freitag,
  et~al.]{goyal2017ssv2}
Raghav Goyal, Samira Ebrahimi~Kahou, Vincent Michalski, Joanna Materzynska,
  Susanne Westphal, Heuna Kim, Valentin Haenel, Ingo Fruend, Peter Yianilos,
  Moritz Mueller-Freitag, et~al.
\newblock The" something something" video database for learning and evaluating
  visual common sense.
\newblock \emph{Proceedings of the IEEE international conference on computer
  vision}, pages 5842--5850, 2017{\natexlab{b}}.

\bibitem[Herman(2010)]{herman2010laboratory}
Louis~M Herman.
\newblock What laboratory research has told us about dolphin cognition.
\newblock \emph{International Journal of Comparative Psychology}, 23\penalty0
  (3), 2010.

\bibitem[Hohwy(2013)]{hohwy2013predictivemind}
Jakob Hohwy.
\newblock \emph{The predictive mind}.
\newblock OUP Oxford, 2013.

\bibitem[Jassim et~al.(2024)Jassim, Holubar, Richter, Wolff, Ohmer, and
  Bruni]{jassim_grasp_2024}
Serwan Jassim, Mario Holubar, Annika Richter, Cornelius Wolff, Xenia Ohmer, and
  Elia Bruni.
\newblock Grasp: A novel benchmark for evaluating language grounding and
  situated physics understanding in multimodal language models.
\newblock In Kate Larson, editor, \emph{Proceedings of the Thirty-Third
  International Joint Conference on Artificial Intelligence, {IJCAI-24}}, pages
  6297--6305, 8 2024.
\newblock \doi{10.24963/ijcai.2024/696}.
\newblock Main Track.

\bibitem[Kang et~al.(2024)Kang, Yue, Lu, Lin, Zhao, Wang, Huang, and
  Feng]{kang2024farvideogenerationworld}
Bingyi Kang, Yang Yue, Rui Lu, Zhijie Lin, Yang Zhao, Kaixin Wang, Gao Huang,
  and Jiashi Feng.
\newblock How far is video generation from world model: A physical law
  perspective.
\newblock \emph{arXiv:2411.02385}, 2024.

\bibitem[Kaplan et~al.(2020)Kaplan, McCandlish, Henighan, Brown, Chess, Child,
  Gray, Radford, Wu, and Amodei]{kaplan2020scaling}
Jared Kaplan, Sam McCandlish, Tom Henighan, Tom~B Brown, Benjamin Chess, Rewon
  Child, Scott Gray, Alec Radford, Jeffrey Wu, and Dario Amodei.
\newblock Scaling laws for neural language models.
\newblock \emph{arXiv:2001.08361}, 2020.

\bibitem[Kay et~al.(2017)Kay, Carreira, Simonyan, Zhang, Hillier,
  Vijayanarasimhan, Viola, Green, Back, Natsev, et~al.]{kay2017kinetics}
Will Kay, Joao Carreira, Karen Simonyan, Brian Zhang, Chloe Hillier, Sudheendra
  Vijayanarasimhan, Fabio Viola, Tim Green, Trevor Back, Paul Natsev, et~al.
\newblock The kinetics human action video dataset.
\newblock \emph{arXiv:1705.06950}, 2017.

\bibitem[Kim and Spelke(1992)]{kim1992gravity}
In~Kyeong Kim and Elizabeth~S Spelke.
\newblock Infants' sensitivity to effects of gravity on visible object motion.
\newblock \emph{Journal of Experimental Psychology: Human Perception and
  Performance}, 18\penalty0 (2):\penalty0 385, 1992.

\bibitem[LeCun(2022)]{lecun2022jepa}
Yann LeCun.
\newblock A path towards autonomous machine intelligence version 0.9. 2,
  2022-06-27.
\newblock \emph{Open Review}, 62\penalty0 (1):\penalty0 1--62, 2022.

\bibitem[Lerer et~al.(2016)Lerer, Gross, and Fergus]{lerer2016learning}
Adam Lerer, Sam Gross, and Rob Fergus.
\newblock Learning physical intuition of block towers by example.
\newblock \emph{International conference on machine learning}, pages 430--438,
  2016.

\bibitem[Long et~al.(2024)Long, Xiang, Stojanov, Sparks, Yin, Keene, Tan, Feng,
  Zhuang, Marchman, Yamins, and Frank]{long_babyview_2024}
Bria Long, Violet Xiang, Stefan Stojanov, Robert~Z. Sparks, Zi~Yin, Grace~E.
  Keene, Alvin W.~M. Tan, Steven~Y. Feng, Chengxu Zhuang, Virginia~A. Marchman,
  Daniel L.~K. Yamins, and Michael~C. Frank.
\newblock The {BabyView} dataset: {High}-resolution egocentric videos of
  infants' and young children's everyday experiences.
\newblock \emph{arXiv:2406.10447}, June 2024.

\bibitem[Loshchilov and Hutter(2019)]{loshchilov2018adamw}
Ilya Loshchilov and Frank Hutter.
\newblock Decoupled weight decay regularization.
\newblock \emph{International Conference on Learning Representations}, 2019.

\bibitem[Margoni et~al.(2024)Margoni, Surian, and
  Baillargeon]{margoni_voe_2024}
Francesco Margoni, Luca Surian, and Renée Baillargeon.
\newblock The violation-of-expectation paradigm: {A} conceptual overview.
\newblock \emph{Psychological Review}, 131\penalty0 (3):\penalty0 716--748,
  April 2024.
\newblock ISSN 1939-1471, 0033-295X.
\newblock \doi{10.1037/rev0000450}.

\bibitem[Mendes et~al.(2007)Mendes, Hanus, and Call]{mendes2007raising}
Natacha Mendes, Daniel Hanus, and Josep Call.
\newblock Raising the level: orangutans use water as a tool.
\newblock \emph{Biology letters}, 3\penalty0 (5):\penalty0 453--455, 2007.

\bibitem[Miech et~al.(2019)Miech, Zhukov, Alayrac, Tapaswi, Laptev, and
  Sivic]{miech2019howto100m}
Antoine Miech, Dimitri Zhukov, Jean-Baptiste Alayrac, Makarand Tapaswi, Ivan
  Laptev, and Josef Sivic.
\newblock Howto100m: Learning a text-video embedding by watching hundred
  million narrated video clips.
\newblock \emph{Proceedings of the IEEE/CVF international conference on
  computer vision}, pages 2630--2640, 2019.

\bibitem[Moravec(1988)]{moravec1988mind}
Hans Moravec.
\newblock Mind children: The future of robot and human intelligence.
\newblock \emph{Harvard UP}, 1988.

\bibitem[Motamed et~al.(2025)Motamed, Culp, Swersky, Jaini, and
  Geirhos]{motamed2025physicsiq}
Saman Motamed, Laura Culp, Kevin Swersky, Priyank Jaini, and Robert Geirhos.
\newblock Do generative video models learn physical principles from watching
  videos?
\newblock \emph{arXiv:2501.09038}, 2025.

\bibitem[OpenAI(2024)]{openai_gpt4_2024}
OpenAI.
\newblock Gpt-4 technical report, 2024.

\bibitem[Piaget(1954)]{piaget1954construction}
Jean Piaget.
\newblock \emph{The Construction of Reality in the Child}.
\newblock Basic Books, 1954.

\bibitem[Piloto et~al.(2022)Piloto, Weinstein, Battaglia, and
  Botvinick]{piloto_intuitive_2022}
Luis~S. Piloto, Ari Weinstein, Peter Battaglia, and Matthew Botvinick.
\newblock Intuitive physics learning in a deep-learning model inspired by
  developmental psychology.
\newblock \emph{Nature Human Behaviour}, 6\penalty0 (9):\penalty0 1257--1267,
  July 2022.
\newblock ISSN 2397-3374.
\newblock \doi{10.1038/s41562-022-01394-8}.

\bibitem[Rao and Ballard(1999)]{rao1999predictivecoding}
Rajesh~PN Rao and Dana~H Ballard.
\newblock Predictive coding in the visual cortex: a functional interpretation
  of some extra-classical receptive-field effects.
\newblock \emph{Nature neuroscience}, 2\penalty0 (1):\penalty0 79--87, 1999.

\bibitem[Reid et~al.(2024)Reid, Savinov, Teplyashin, Lepikhin, Lillicrap,
  Alayrac, Soricut, Lazaridou, Firat, Schrittwieser, et~al.]{reid2024gemini}
Machel Reid, Nikolay Savinov, Denis Teplyashin, Dmitry Lepikhin, Timothy
  Lillicrap, Jean-baptiste Alayrac, Radu Soricut, Angeliki Lazaridou, Orhan
  Firat, Julian Schrittwieser, et~al.
\newblock Gemini 1.5: Unlocking multimodal understanding across millions of
  tokens of context.
\newblock \emph{arXiv:2403.05530}, 2024.

\bibitem[Riochet et~al.(2020)Riochet, Sivic, Laptev, and
  Dupoux]{riochet2020occlusion}
Ronan Riochet, Josef Sivic, Ivan Laptev, and Emmanuel Dupoux.
\newblock Occlusion resistant learning of intuitive physics from videos.
\newblock \emph{arXiv:2005.00069}, 2020.

\bibitem[Riochet et~al.(2022)Riochet, Castro, Bernard, Lerer, Fergus, Izard,
  and Dupoux]{riochet_intphys_2022}
Ronan Riochet, Mario~Ynocente Castro, Mathieu Bernard, Adam Lerer, Rob Fergus,
  Véronique Izard, and Emmanuel Dupoux.
\newblock {IntPhys} 2019: {A} {Benchmark} for {Visual} {Intuitive} {Physics}
  {Understanding}.
\newblock \emph{IEEE Transactions on Pattern Analysis and Machine
  Intelligence}, 44\penalty0 (9):\penalty0 5016--5025, September 2022.
\newblock ISSN 1939-3539.
\newblock \doi{10.1109/TPAMI.2021.3083839}.

\bibitem[Singer and Henderson(2015)]{singer2015object}
Rebecca Singer and Elizabeth Henderson.
\newblock Object permanence in marine mammals using the violation of
  expectation procedure.
\newblock \emph{Behavioural Processes}, 112:\penalty0 108--113, 2015.

\bibitem[Smith et~al.(2019)Smith, Mei, Yao, Wu, Spelke, Tenenbaum, and
  Ullman]{smith_adept_2019}
Kevin Smith, Lingjie Mei, Shunyu Yao, Jiajun Wu, Elizabeth Spelke, Josh
  Tenenbaum, and Tomer Ullman.
\newblock Modeling expectation violation in intuitive physics with coarse
  probabilistic object representations.
\newblock \emph{Advances in neural information processing systems}, 32, 2019.

\bibitem[Spelke(1985)]{spelke1985voe_looking}
Elizabeth~S Spelke.
\newblock Preferential-looking methods as tools for the study of cognition in
  infancy.
\newblock In Gilbert Gottlieb and Norman~A. Krasnegor, editors,
  \emph{Measurement of audition and vision in the first year of postnatal life:
  A methodological overview}, pages 323--363. Ablex Publishing, 1985.

\bibitem[Spelke(2000)]{spelke_core_2000}
Elizabeth~S. Spelke.
\newblock Core knowledge.
\newblock \emph{American Psychologist}, 55\penalty0 (11):\penalty0 1233--1243,
  2000.
\newblock ISSN 1935-990X.
\newblock \doi{10.1037/0003-066X.55.11.1233}.
\newblock Place: US Publisher: American Psychological Association.

\bibitem[Spelke and Kinzler(2007)]{spelke_core_knowledge_2007}
Elizabeth~S. Spelke and Katherine~D. Kinzler.
\newblock Core knowledge.
\newblock \emph{Developmental Science}, 10\penalty0 (1):\penalty0 89--96,
  January 2007.
\newblock ISSN 1363-755X, 1467-7687.
\newblock \doi{10.1111/j.1467-7687.2007.00569.x}.

\bibitem[Spelke et~al.(1992)Spelke, Breinlinger, Macomber, and
  Jacobson]{spelke_origins_1992}
Elizabeth~S. Spelke, Karen Breinlinger, Janet Macomber, and Kristen Jacobson.
\newblock Origins of knowledge.
\newblock \emph{Psychological Review}, 99\penalty0 (4):\penalty0 605--632,
  1992.
\newblock ISSN 1939-1471, 0033-295X.
\newblock \doi{10.1037/0033-295X.99.4.605}.

\bibitem[Spelke et~al.(1995)Spelke, Kestenbaum, Simons, and
  Wein]{spelke_spatiotemporal_1995}
Elizabeth~S. Spelke, Roberta Kestenbaum, Daniel~J. Simons, and Debra Wein.
\newblock Spatiotemporal continuity, smoothness of motion and object identity
  in infancy.
\newblock \emph{British Journal of Developmental Psychology}, 13\penalty0
  (2):\penalty0 113--142, June 1995.
\newblock ISSN 0261-510X, 2044-835X.
\newblock \doi{10.1111/j.2044-835X.1995.tb00669.x}.

\bibitem[Su et~al.(2021)Su, Lu, Pan, Wen, and Liu]{su2021rope}
Jianlin Su, Yu~Lu, Shengfeng Pan, Bo~Wen, and Yunfeng Liu.
\newblock Roformer: enhanced transformer with rotary position embedding. corr
  abs/2104.09864 (2021).
\newblock \emph{arXiv:2104.09864}, 2021.

\bibitem[Sullivan et~al.(2021)Sullivan, Mei, Perfors, Wojcik, and
  Frank]{sullivan2021saycam}
Jessica Sullivan, Michelle Mei, Andrew Perfors, Erica Wojcik, and Michael~C
  Frank.
\newblock Saycam: A large, longitudinal audiovisual dataset recorded from the
  infant’s perspective.
\newblock \emph{Open mind}, 5:\penalty0 20--29, 2021.

\bibitem[Taylor et~al.(2012)Taylor, Miller, and Gray]{taylor2012new}
Alex~H Taylor, Rachael Miller, and Russell~D Gray.
\newblock New caledonian crows reason about hidden causal agents.
\newblock \emph{Proceedings of the National Academy of Sciences}, 109\penalty0
  (40):\penalty0 16389--16391, 2012.

\bibitem[Ullman et~al.(2017)Ullman, Spelke, Battaglia, and
  Tenenbaum]{ullman2017mind}
Tomer~D Ullman, Elizabeth Spelke, Peter Battaglia, and Joshua~B Tenenbaum.
\newblock Mind games: Game engines as an architecture for intuitive physics.
\newblock \emph{Trends in cognitive sciences}, 21\penalty0 (9):\penalty0
  649--665, 2017.

\bibitem[Vallortigara(2012)]{vallortigara2012core}
Giorgio Vallortigara.
\newblock Core knowledge of object, number, and geometry: A comparative and
  neural approach.
\newblock \emph{Cognitive neuropsychology}, 29\penalty0 (1-2):\penalty0
  213--236, 2012.

\bibitem[Wang et~al.(2023)Wang, Huang, Zhao, Tong, He, Wang, Wang, and
  Qiao]{wang_videomaev2_2023}
Limin Wang, Bingkun Huang, Zhiyu Zhao, Zhan Tong, Yinan He, Yi~Wang, Yali Wang,
  and Yu~Qiao.
\newblock Videomae v2: Scaling video masked autoencoders with dual masking.
\newblock \emph{Proceedings of the IEEE/CVF Conference on Computer Vision and
  Pattern Recognition (CVPR)}, pages 14549--14560, June 2023.

\bibitem[Wang et~al.(2024)Wang, Bai, Tan, Wang, Fan, Bai, Chen, Liu, Wang, Ge,
  Fan, Dang, Du, Ren, Men, Liu, Zhou, Zhou, and Lin]{wang_2024_qwen}
Peng Wang, Shuai Bai, Sinan Tan, Shijie Wang, Zhihao Fan, Jinze Bai, Keqin
  Chen, Xuejing Liu, Jialin Wang, Wenbin Ge, Yang Fan, Kai Dang, Mengfei Du,
  Xuancheng Ren, Rui Men, Dayiheng Liu, Chang Zhou, Jingren Zhou, and Junyang
  Lin.
\newblock Qwen2-vl: Enhancing vision-language model's perception of the world
  at any resolution.
\newblock \emph{arXiv:2409.1219}, 2024.

\bibitem[Watters et~al.(2017)Watters, Tacchetti, Weber, Pascanu, Battaglia, and
  Zoran]{watters_visual_interaction_networks_2017}
Nicholas Watters, Andrea Tacchetti, Theophane Weber, Razvan Pascanu, Peter
  Battaglia, and Daniel Zoran.
\newblock Visual {Interaction} {Networks}.
\newblock \emph{arXiv:1706.01433}, June 2017.

\bibitem[Wei et~al.(2023)Wei, Wang, Schuurmans, Bosma, Ichter, Xia, Chi, Le,
  and Zhou]{wei2023cot}
Jason Wei, Xuezhi Wang, Dale Schuurmans, Maarten Bosma, Brian Ichter, Fei Xia,
  Ed~Chi, Quoc Le, and Denny Zhou.
\newblock Chain-of-thought prompting elicits reasoning in large language
  models.
\newblock \emph{arXiv:2201.11903}, 2023.

\bibitem[Weihs et~al.(2022)Weihs, Yuile, Baillargeon, Fisher, Marcus, Mottaghi,
  and Kembhavi]{Weihs_InfLevel_2022}
Luca Weihs, Amanda~Rose Yuile, Ren\'{e}e Baillargeon, Cynthia Fisher, Gary
  Marcus, Roozbeh Mottaghi, and Aniruddha Kembhavi.
\newblock Benchmarking progress to infant-level physical reasoning in ai.
\newblock \emph{TMLR}, 2022.

\bibitem[Wilcox(1999)]{wilcox1999constancy}
Teresa Wilcox.
\newblock Object individuation: Infants’ use of shape, size, pattern, and
  color.
\newblock \emph{Cognition}, 72\penalty0 (2):\penalty0 125--166, 1999.

\bibitem[Wilcox and Chapa(2004)]{wilcox2004constancy}
Teresa Wilcox and Catherine Chapa.
\newblock Priming infants to attend to color and pattern information in an
  individuation task.
\newblock \emph{Cognition}, 90\penalty0 (3):\penalty0 265--302, 2004.

\bibitem[Wood(2013)]{wood2013newborn}
Justin~N Wood.
\newblock Newborn chickens generate invariant object representations at the
  onset of visual object experience.
\newblock \emph{Proceedings of the National Academy of Sciences}, 110\penalty0
  (34):\penalty0 14000--14005, 2013.

\end{thebibliography}

\clearpage
\newpage
\beginappendix

\renewcommand{\thefigure}{S\arabic{figure}}
\renewcommand{\thetable}{S\arabic{table}}
\renewcommand{\theequation}{S\arabic{equation}}
\setcounter{figure}{0}
\setcounter{table}{0}
\setcounter{equation}{0}

\section{Materials and Methods}

\subsection{Unsupervised pretraining of V-JEPA~\label{sec:method}}

V-JEPA~\citep{bardes_vjepa_2024} is composed of multiple components. First, a context encoder $f_\theta$ whose goal is to output abstract representations of a corrupted video. A target encoder $f_{\theta^{EMA}}$ is used to encode the full video and produce targets for the predictor. The weights of the target encoder $\theta^{EMA}$ are an exponential moving average of the weights of the context encoder $\theta$. For an exponential moving average parameter $\alpha \in [0,1]$ and at iteration $t$ during training, we get the update rule of $\theta^{EMA}_{t+1} = (1-\alpha)\theta_{t} + \alpha \theta^{EMA}_{t}$.
Finally, the predictor $p_\phi$ is used to predict the uncorrupted representations from the corrupted ones.
During training, we start from a video $V$, which is corrupted into $V_C$, by masking random blocks in the video. The target thus becomes the complementary $\bar{V_C}$. The training objective is thus to predict the representations of $\bar{V_C}$ from $V_C$ by minimizing the following objective:
\begin{equation}
 \|p_\phi\left(f_\theta\left(V_C\right)\right) - f_{\theta^{EMA}}(\bar{V_C}) \|_1 .   
\end{equation}

While at training time the corruption used is the removal of spatio-temporal blocks, we can see that if we instead use the first $C$ frames to predict the rest of the video, this objective turns into a measure of error for the prediction of the future.

\subsection{Pretraining Data}

For the pretraining of V-JEPA we rely on multiple data sources. The original data mix that was used is VideoMix2M \citep{bardes_vjepa_2024}, which is the concatenation of three datasets: Kinetics710~\citep{kay2017kinetics}, SomethingSomething-v2~\citep{goyal2017ssv2} and HowTo100M~\citep{miech2019howto100m}. Kinetics710 consists of around 650k videos spanning 710 action classes (e.g. kayaking, ironing, etc.), each lasting around 10 seconds.\\ SomethingSomething-v2 is focused more on motions, where we find classes such as "Uncovering something" or "Throwing something". It consists of around 200k clips that last a few seconds on average. HowTo100M is a much larger dataset, containing around 1.2M videos lasting 6m30s on average, for a total of around 15 years of unique video data. Here, the individuals are not curated as precisely as Kinetics or SomethingSomething, yielding a more "in the wild" data source.

As discussed in the main text, most of our experiments are done with only HowTo100m, which exhibits the highest performance and demonstrates how V-JEPA can leverage uncurated data sources.

\subsection{V-JEPA pretraining hyperparameters}

\begin{table}
    \centering
        \caption{\textbf{Pretraining hyper-parameters for V-JEPA.} Table structure and values identical to the original V-JEPA paper~\citep{bardes_vjepa_2024}, apart from positional embedding where we rely on RoPE~\citep{su2021rope}.}
    \label{tab:app_hyper}
    {\fontsize{10pt}{10pt}\selectfont
    \setlength{\tabcolsep}{2pt}
    \begin{tabular}{@{} l c c c @{}}
        \hline
        Hyper-parameter & ViT-B/16 & ViT-L/16 &
        ViT-H/16 \\
        \hline \textit{positional embeddings}\\
        Type & RoPE & RoPE & RoPE\\
        theta & 10000 & 10000 & 10000 \\
        
        \hline \textit{data} \\
        resolution & 224 & 224 & 224 \\
        num\_frames & 16 & 16 & 16 \\
        framerate & 5.33 fps & 5.33 fps & 5.33 fps \\
        horizontal\_flip & true & true & true \\
        random\_resize\_scale & (0.3, 1.0) & (0.3, 1.0) & (0.3, 1.0)\\
        random\_resize\_aspect\_ratio & (0.75, 1.35) & (0.75, 1.35) & (0.75, 1.35)\\
        
        \hline \textit{masking} \\
        block\_aspect\_ratio & (0.75, 1.5) & (0.75, 1.5) & (0.75,1.5)\\
        shortrange\_mask\_num\_blocks & 8 & 8 & 8\\
        shortrange\_mask\_spatial\_scale & 0.15 & 0.15 & 0.15\\
        longrange\_mask\_num\_blocks & 2 & 2 & 2\\
        longrange\_mask\_spatial\_scale & 0.7 & 0.7 & 0.7\\
        
        \hline \textit{optimization} \\
        optimizer & AdamW & AdamW & AdamW\\
        batch\_size & 3072 & 3072 & 3072\\
        total\_number\_of\_iterations & 90000 & 90000 & 90000\\
        scheduler & Linear + Cosine & Linear + Cosine & Linear + Cosine \\
        warmup\_iterations & 12000 & 12000 & 12000 \\
        learning\_rate & 6.25$\times10^{-4}$ &6.25$\times10^{-4}$ & 6.25$\times10^{-4}$ \\
        start\_lr & 2$\times10^{-4}$ & 2$\times10^{-4}$ & 2$\times10^{-4}$ \\
        final\_lr & 1$\times10^{-6}$ & 1$\times10^{-6}$ & 1$\times10^{-6}$ \\
        start\_momentum & 0.998 & 0.998 & 0.998\\
        final\_momentum & 1.0 & 1.0 & 1.0 \\
        start\_weight\_decay & 0.04 & 0.04 & 0.04 \\
        final\_weight\_decay & 0.4 & 0.4 & 0.4\\
        scheduler\_scale\_factor & 1.25 & 1.25 & 1.25 \\

        \hline \textit{architecture} \\
        patch\_size & 16 & 16 & 16\\
        tubelet\_size & 2 & 2 & 2 \\
        pred\_depth & 12 & 12 & 12\\
        pred\_embed\_dim & 384 & 384 & 384 \\
        
        \hline \textit{hardware} \\
        dtype & bfloat16 & bfloat16 & bfloat16\\
        accelerator & A100 80G & A100 80G & A100 80G \\
        \hline
    \end{tabular}}

\end{table}

Throughout our experiments, we use the same set of hyperparameters across models, only varying the ablated component. We provide a summary of the hyperparameters in Table~\ref{tab:app_hyper}. We rely on the same training protocol as in the original V-JEPA paper~\citep{bardes_vjepa_2024}, but use RoPE~\citep{su2021rope} to encode positional information instead of absolute positional embeddings. To use RoPE on 3D data (height$\times$width$\times$time) we split the feature dimensions in 3 and use each third for a spatiotemporal-dimension.

Here, we expand on three core elements: architecture, optimization and masking.

\textbf{Architecture.} We use the Vision Transformer (ViT)~\citep{dosovitskiy2021an} for the context encoder and target encoder. All encoders are trained to take in a clip of maximum 3 seconds over 16 frames (5.33 fps), at a resolution of $224 \times 224$. The video clips are flattened in a sequence of non-overlapping patches of shape $16 \times 16 \times 2$. The predictor also uses a ViT inspired architecture, consisting of 12 blocks with a smaller embedding dimension of 384.

\textbf{Optimization.} We rely on the AdamW~\citep{loshchilov2018adamw} optimizer to train the context encoder and predictor. For all experiments, we use a batch size of 3072, where we train the models for 90000 iterations. This corresponds to a total of around 26 years of (non-unique) video. The learning rate is increased linearly from $2\times10^{-4}$ to $6.25\times10^{-4}$  over the first 12000 iterations. The learning rate is then decayed to $1\times10^{-6}$ following a cosine schedule during the remaining iterations. We stretch the schedule by a factor of $1.25$, meaning that the schedule lasts for 112500 iterations of which we only perform 90000. This avoids a decay of performance at the end of training when the learning rate becomes too small, which would lead the context and target encoder to collapse partially.

\textbf{Masking.} In our experiments, we rely on a few masking strategies that we describe precisely here.
\begin{itemize}
    \item Block masking. We mask the union of 8 blocks with a spatial scale of 0.15, as well as 2 blocks with scale 0.7. For all blocks, we use an aspect ratio sampled uniformly between 0.75 and 1.5. This strategy is used unless specified otherwise.
    \item Causal Block masking. This strategy is identical to Block masking, where we additionally fully mask the last 4 frames of the video clips.
    \item Random masking. This strategy randomly masks 90\% of all patches in the video clips, following a uniform distribution.
\end{itemize}

\subsection{Evaluation Data}

\begin{table}[!h]
    \centering
    \caption{Summary of datasets used for evaluation. IntPhys, GRASP and InfLevel-lab provide qualitatively different data sources to perform a more holistic evaluation of models.}
    \begin{tabular}{lcccc}
        \hline    
        Dataset &  Realistic & Diverse scenes & Size & Number of Properties  \\
        \hline
        IntPhys & No & Yes & $\sim$ 360 & 3 \\
        GRASP & No & No & $\sim$ 4000 & 10 \\
        InfLevel-lab & Yes & No & $\sim$ 4000 & 3 \\

        \hline
    \end{tabular}
    
    \label{tab:eval-datasets}
\end{table}

To provide a more general assessment of the models considered, we focus on multiple data sources for evaluation, for which we summarize the main characteristics in Table~\ref{tab:eval-datasets}. IntPhys~\citep{riochet_intphys_2022} is the most carefully curated data source, with pairs of videos being aligned at the pixel level thanks to the use of a simulator and each frame being individually stored (avoiding compression artifacts). Due to its formulation as a challenge with a private test set, we rely on the smaller "dev" subset of the data, which has publicly available labels. Nevertheless, for each pair of videos, the number of objects, occluders, and texture/shape/color of objects are randomized, which ensures that the model performs well in diverse environments. 

GRASP~\citep{jassim_grasp_2024} is similar to IntPhys in the sense that it also uses simulated data, but covers a wider range of properties (10 compared to 3) with more total videos. A caveat to its use for our study is that it was originally designed to evaluate models on a single video rather than a pair. As such, even if the videos are paired in practice, we found some issues regarding the performance of untrained networks where they could latch on spurious features and achieve high accuracy. Some videos in GRASP can be attributed to multiple properties, so in practice, we consider that they belong to all properties when presenting the results (e.g., a video belonging to gravity and support will be counted for both gravity and support separately).

InfLevel-lab~\citep{Weihs_InfLevel_2022} gives us a source of natural videos where the manipulations are remarkably paired. The only visual differences between videos inside a given pair are in the lighting of the scene, to which models should be robust. Here, three properties are tested; however, for two of them (solidity and gravity), the model needs to first see a contextualizing event where the objects used during the manipulation are shown. Without seeing this pretext video, the task becomes impossible. Inflevel-lab thus requires more memory and adaptability than the tested models have. We further rename the 'continuity' property as 'object permanence' for consistency with other datasets. The difference between the two is subtle, but the experimental setup in InfLevel-lab is closer to the one of object permanence in IntPhys and GRASP.

We emphasize that none of these datasets are seen during training and are only used for evaluation purposes where the networks are frozen, making all of these datasets out of distribution.

\subsection{Properties\label{sec:properties}}

We provide brief descriptions of the intuitive physical properties considered in our work:

\noindent\textbf{Object permanence~\citep{baillargeon_permanence_1991}.} Objects do not spontaneously materialize or vanish out of thin air. Objects also keep existing when occluded.

\noindent\textbf{Continuity~\citep{spelke_origins_1992}.} Objects follow a continuous path and do not teleport in space or time.
This concept is closely linked to object permanence, but leads to more subtle differences in experimental setups.

\noindent\textbf{Shape and color constancy~\citep{wilcox1999constancy,wilcox2004constancy}.} Objects do not spontaneously change color or shape.

\noindent\textbf{Gravity~\citep{kim1992gravity}.} Objects fall down without support under them.

\noindent\textbf{Support~\citep{baillargeon_support_1990,baillargeon_support_1992}.} Objects are stable when positioned on a platform, but become unstable/fall when unsupported. This is closely related to gravity, where the main difference lies in the precise experimental setup. For example, an object can be pushed off of a platform to test support, and an object can just be dropped in the air to test gravity.

\noindent\textbf{Solidity~\citep{spelke_origins_1992}.} Objects cannot overlap or pass through each other. When tested behind an occluder, this property shares similarities with continuity, where an object should also not teleport across another object.

\noindent\textbf{Inertia~\citep{spelke_origins_1992}.} Inanimate objects do not spontaneously alter their motion, such as a change in direction.

\noindent\textbf{Collision~\citep{baillargeon1995collision}.} Objects do not stay still when hit by another similar moving object.

\noindent For exact experimental setups, we refer the reader to the original datasets~\citep{riochet_intphys_2022,jassim_grasp_2024,Weihs_InfLevel_2022},

\subsection{Baselines}

\textbf{Pixel prediction.} For our pixel prediction baseline, we use VideoMAEv2~\citep{wang_videomaev2_2023} which is trained similarly to V-JEPA. However, instead of predicting missing parts of the video in latent space, this objective is solved in normalized pixel space. Each patch of 16x16x2 pixels is first normalized before being used as a target. This makes VideoMAEv2 a good comparison with V-JEPA due to the similarity in implementation details while being a fundamentally different framework.

\textbf{Multimodal Large Language Models.} for this baseline, we choose to use Qwen2-VL~\citep{wang_2024_qwen}, one of the best open source Multimodal LLM that can handle video inputs at the time of writing, as well as Gemini 1.5 pro, a proprietary Multimodal LLM which shines at video understanding. Using Qwen2-VL allows us to have full control over how the video is processed (e.g., some proprietary models~\citep{reid2024gemini} downsample the video to 1 fps) and provide easily reproducible results.

\subsection{Evaluation Protocol\label{sec:surprise}}

\begin{figure}
    \centering
    \includegraphics[width=1\linewidth]{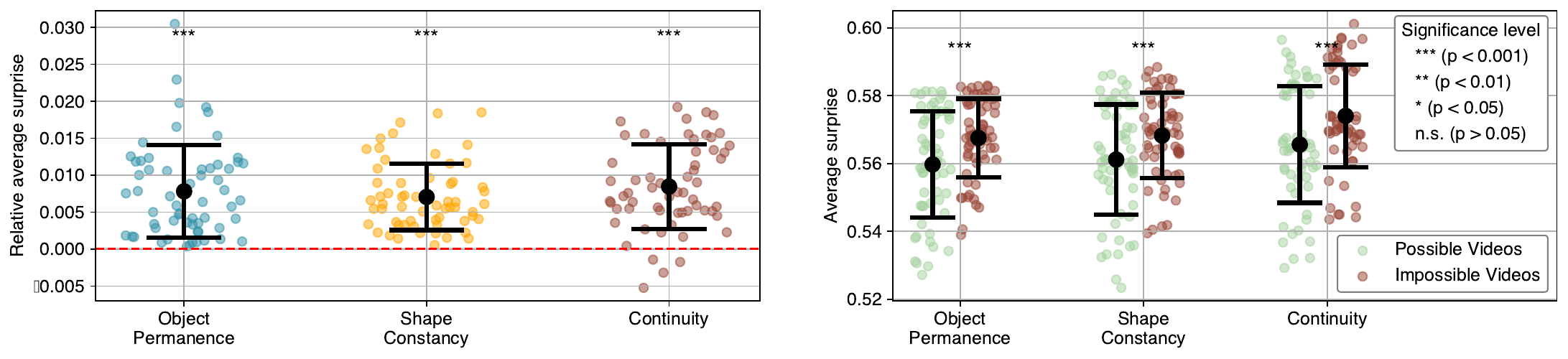}
    \includegraphics[width=1\linewidth]{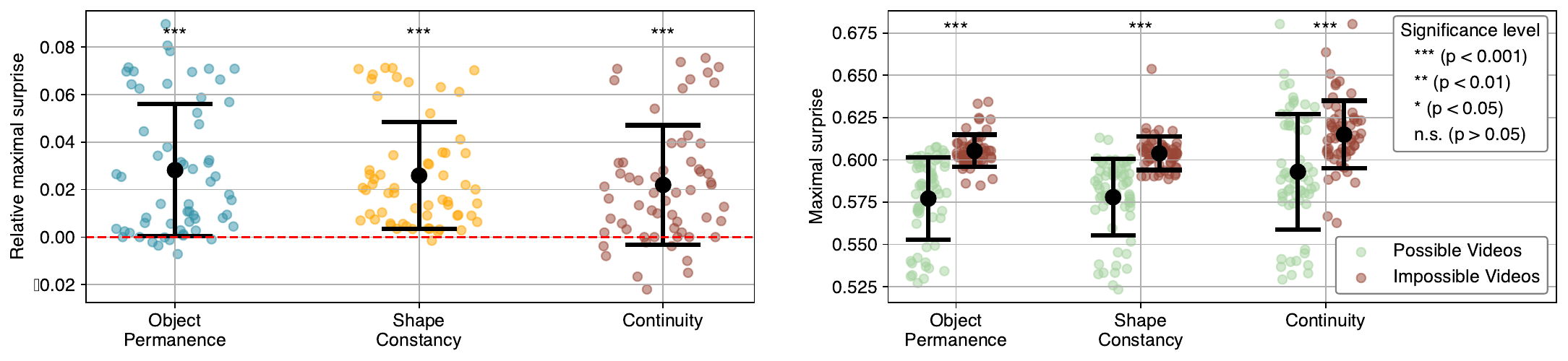}
    \caption{\textbf{Different surprise measures are better suited for different tasks.} Focusing on IntPhys, we find that looking at the average surprise over a video leads to better performance when comparing pairs of videos. A one-sample t-test was performed to see if the relative surprises are greater than zero \textbf{(left)}. However, when looking at individual videos' surprise, choosing the maximum surprise over a video leads to a better separation between possible and impossible videos. A two-sample t-test was performed to see if impossible videos have higher surprise than possible ones. \textbf{(rigt)}.}
    \label{fig:ablation_surprise}
\end{figure}

\textbf{Prediction based methods.} Both pixel and latent prediction methods can be evaluated in the same way, with the only difference being how the target of the prediction is encoded. For V-JEPA, we use abstract representations of the future obtained by encoding the video and then only keeping the future frames, while for VideoMAEv2 the target is simply the normalized future of the video.
Considering a video $V$ with frames $1,...,T$, a context encoder $f_\theta$ handling $C$ frames, a target encoder $g_\psi$ producing the groundtruth $M$ future frames from the video, and a predictor predicting $M$ frames in the future, we can measure surprise at time $t$ as
\begin{equation}
    S_t = \| p_\phi\left(f_\theta\left(V_{t:t+C}\right) \right) - g_\psi\left(V_{t:t+C+M}\right) \|_1 .
\end{equation}
This surprise can then be computed over a full video to obtain a global surprise score:
\begin{equation}
    \texttt{AvgSurprise} = \frac{1}{T} \sum_{t \in \{1,1+s,...,T-(C+M)\}} S_t \quad\textbf{or}\quad     \texttt{MaxSurprise} = \max_{t \in \{1,1+s,...,T-(C+M)\}}S_t .
\end{equation}
Where $s$ is a stride parameter which helps reduce the amount of compute used. In practice, we use $s = 2$, which corresponds to predicting starting from frames $1,3,5,$ etc.

For each property, we select the value of $C$ ($C + M$ being fixed) which maximizes performance to account for different constraints on memory for various tasks. On IntPhys, we are able to get rid of this sweep on context lengths by computing the minimal surprise over all context lengths $C$ for each starting frame $t$.

By comparing the surprise score between possible and impossible videos, we can thus measure whether or not the model has understood the intrinsic physical property.
Using an average surprise score~\citep{smith_adept_2019,piloto_intuitive_2022} is ideal for comparing similar videos, but a maximum surprise score can be used on unique videos by eliminating the surprise contribution coming from the complexity of the scene.
Relative surprise~\citep{piloto_intuitive_2022,smith_adept_2019,riochet_intphys_2022}, which looks at the difference between surprise on the impossible and possible videos, is commonly used as it allows to precisely measure the effect of the physics breaking event. Focusing on IntPhys, we performed a single-tail one sample t-test to assess whether the model exhibits a relative surprise greater than zero. Using the average surprise of a video, we find that for all properties, V-JEPA produces a relative surprise over zero: Object Permanence: M=7.8e-03, SD=6.3e-03 (t(59.0) = 9.7, $p$ = $4.64\times10^{-14}$);Shape Constancy: M=7.1e-03, SD=4.5e-03 (t(59.0) = 12.2, $p$ = $5.29\times10^{-18}$);Continuity: M=8.5e-03, SD=5.7e-03 (t(59.0) = 11.5, $p$ = $6.03\times10^{-17}$). These results can be visualized in the left column in Figure~\ref{fig:ablation_surprise}.
Using the maximum surprise, we find that for all properties, V-JEPA produces a relative surprise over zero: Object Permanence: M=8.5e-03, SD=5.7e-03 (t(59.0) = 7.9, $p$ = $4.54\times10^{-11}$);Shape Constancy: M=8.5e-03, SD=5.7e-03 (t(59.0) = 8.9, $p$ = $7.51\times10^{-13}$);Continuity: M=8.5e-03, SD=5.7e-03 (t(59.0) = 6.8, $p$ = $3.19\times10^{-9}$).'
For this pairwise classification task, we find that using the average surprise over a video performs better, however both strategies provide high performance.

A harder task, but closer to reality, is to look at the surprises of possible and impossible videos individually rather than in pairs. Being able to separate possible and impossible videos without pairs is significantly harder~\citep{riochet_intphys_2022,riochet2020occlusion} and requires a deeper understanding of the tested properties. Focusing on IntPhys, we perform a one-tailed two-sample Welch's t-test to assess whether the impossible videos have higher average surprise than possible ones.

Using the average surprise over a video, we find that for all properties, impossible videos had a higher surprise than possible ones on average: Object Permanence: M=5.7e-01, SD=1.2e-02 vs M=5.60e-01, SD=1.57e-02 (t(108.5) = 3.1, $p$ = $1.23\times10^{-3}$);Shape Constancy: M=5.7e-01, SD=1.3e-02 vs M=5.61e-01, SD=1.62e-02 (t(111.2) = 2.7, $p$ = $4.46\times10^{-3}$);Continuity: M=5.7e-01, SD=1.5e-02 vs M=5.66e-01, SD=1.72e-02 (t(116.3) = 2.9, $p$ = $2.52\times10^{-3}$).
Using the maximum surprise over a video, we find that for all properties, impossible videos had a higher surprise than possible ones on average: Object Permanence: M=6.1e-01, SD=9.5e-03 vs M=5.77e-01, SD=2.43e-02 (t(76.7) = 8.4, $p$ = $1.03\times10^{-12}$);Shape Constancy: M=6.0e-01, SD=9.9e-03 vs M=5.78e-01, SD=2.26e-02 (t(80.6) = 8.1, $p$ = $2.06\times10^{-12}$);Continuity: M=6.1e-01, SD=2.0e-02 vs M=5.93e-01, SD=3.41e-02 (t(95.3) = 4.3, $p$ = $2.01\times10^{-5}$). These results can be visualized in the right column of Figure~\ref{fig:ablation_surprise}.
For this task, using the maximum surprise of a video rather than the average is more desirable as it should lead to measures that only focus on the most surprising event, and are thus agnostic to other properties of the video.\\

\textbf{Multimodal LLM.} Due to the output of the models being text only, the most direct approach is simply to ask the model which video is impossible in a pair. We use the following prompt, inspired by the ones used in GRASP~\citep{jassim_grasp_2024} for single video classification:\\
"Video 1: \texttt{<video\_1>}, Video 2: \texttt{<video\_2>}.
You are seeing a pair of videos, Video 1 and Video 2. They were both generated in a simulator, so ignore the quality of the videos. Exactly one of the two videos has an event which breaks the laws of physics. Given how objects behave on Earth, which one is it ? End your answer with the video name."\\
Where \texttt{<video\_1>} and \texttt{<video\_2>} are replaced by the pair of video. We shuffle the order of the videos to avoid any bias where the model may prefer the first/second video.

Asking the model to end its answer by the video name allows us to convert the output to predicted videos easily. We experimented with different other strategies such as 0-shot chain of thought~\citep{wei2023cot}, or using more detailed prompts, but did not find any qualitative or quantitative difference in behavior. For all models we set the sampling temperature to 0. When a model refuses to answer the question, e.g. answering "Both videos are plausible" we count this as an error. Qwen2-VL never fell in this cases but Gemini 1.5 pro did around 10-15\% of the time.

Since we have access to the model for Qwen2-VL, and its output is a probability distribution over possible tokens, we can also look at the probabilities assigned to the choice of video, i.e. "1" or "2" at the end of the answer.
We thus computed normalized probabilities for each video as 
\begin{equation}  
P = \frac{P("1")}{P("1") + P("2")} \quad or \quad \frac{P("2")}{P("1") + P("2")}
\end{equation}

\begin{figure}
    \centering
    \includegraphics[width=1\linewidth]{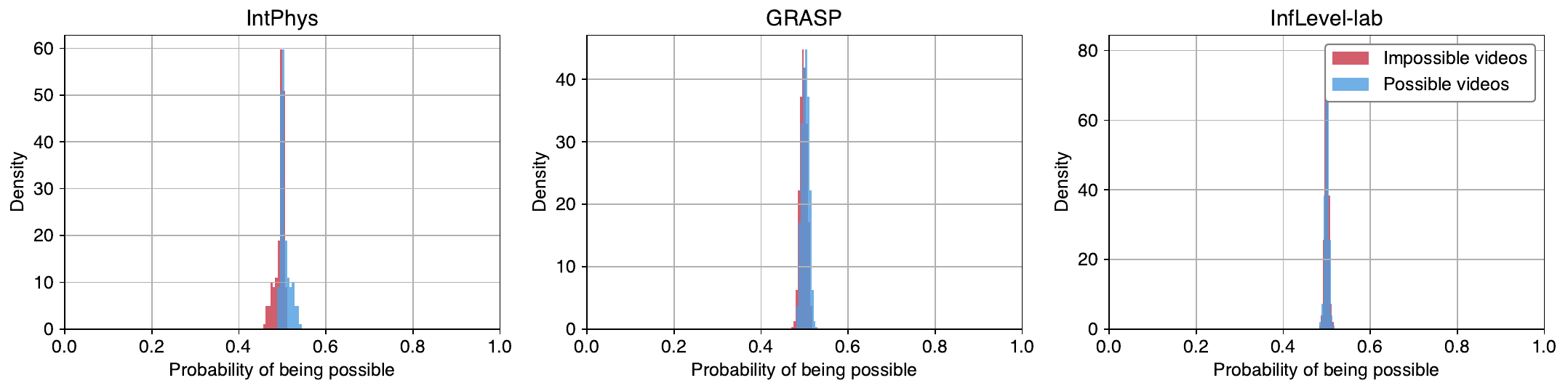}
    \caption{\textbf{Normalized probabilities output by Qwen2-VL-72B.} When presented with a pair of videos, we find that the model outputs similar probabilities for possible and impossible videos.}
    \label{fig:probas_qwen}
\end{figure}

This gives us a more granular surprise measure, although we find that the probability is often around 0.5, meaning that the model predicts almost as if it were a coin toss. This can be seen in Figure~\ref{fig:probas_qwen} for Qwen2-VL-72B.

\subsection{Evaluation hyperparameters\label{sec:hyperparams}}

For every method, we use the following hyperparameters per dataset:
\begin{itemize}
    \item \textbf{IntPhys:} Frame skip in [2,5,10];  Window size ($C+M$) in [16,32]; Context lengths [2,4,6,8,10]$\times (C+M)/16$
     \item \textbf{GRASP:} Frame skip in [2,5,10];  Window size ($C+M$) in [16,32]; Context lengths [2,4,6,8,10]$\times (C+M)/16$
      \item \textbf{InfLevel-lab:} Frame skip in [5,10,20] for V-JEPA and VideoMAEv2, [5,10,20,30] for Qwen and Gemini; Window size ($C+M$) in [16,32]; Context lengths [2,4,6,8,10]$\times (C+M)/16$; 
\end{itemize}

\begin{table}[!h]
\centering
\caption{\textbf{Evaluation hyperparameters}.}

\label{tab:hp-eval}

\begin{tabular}{llcccc}
\hline
Dataset & Method & Frame skip & FPS & Window size & Window Stride \\
\hline
IntPhys& V-JEPA & 2 & 7.5 & 16 & 2 \\
                        & VideoMAEv2 & 2 & 7.5 & 16 & 2 \\
                        & Qwen-2-VL-72b & 5 & 3 & All & N/A \\
                        & Gemini-1.5-pro & 2 & 7.5 & All & N/A \\
\hline
GRASP& V-JEPA & 10 & 5 & 16 & 2 \\
                        & VideoMAEv2 & 10 & 5 & 16 & 2 \\
                        & Qwen-2-VL-72b & 10 & 5 & All & N/A\\
                        & Gemini-1.5-pro & 10 & 5 & All & N/A \\
\hline
InfLevel-lab& V-JEPA & 5 & 6 & 32 & 2 \\
                        & VideoMAEv2 & 10 & 3 & 16 & 2 \\
                        & Qwen-2-VL-72b & 20 & 1.5 & All & N/A\\
                        & Gemini-1.5-pro & 30 & 1 & All & N/A \\

\hline
\end{tabular}
\end{table}

For all properties, we choose the context size which gives us the best performance. This means that on a given dataset, different properties may use different optimal context sizes. For IntPhys, we find that using the minimal surprise over all windows for each start frame can be used as it removes one hyperparameter to optimize and helps filter surprise spikes coming from possible events. For example, an object entering the scene is hard to predict and leads to a surprise spike, but this filtering enables us to remove it.

\section{Choice of prediction hyperparameters and influence on 0-shot performance\label{sec:context}}

As described in materials and methods, a choice of prediction-related hyperparameters must be made to evaluate prediction-based models, such as V-JEPA or VideoMAEv2. Due to the 0-shot nature of the evaluation, the models have no a priori calibration on the task they are solving.

We thus need to find a way to select hyperparameters, especially the context length for the prediction. This directly dictates how far back in the past models can look, and how far in the future they are predicting.

\begin{figure}[!p]
    \centering
    \includegraphics[width=0.8\linewidth]{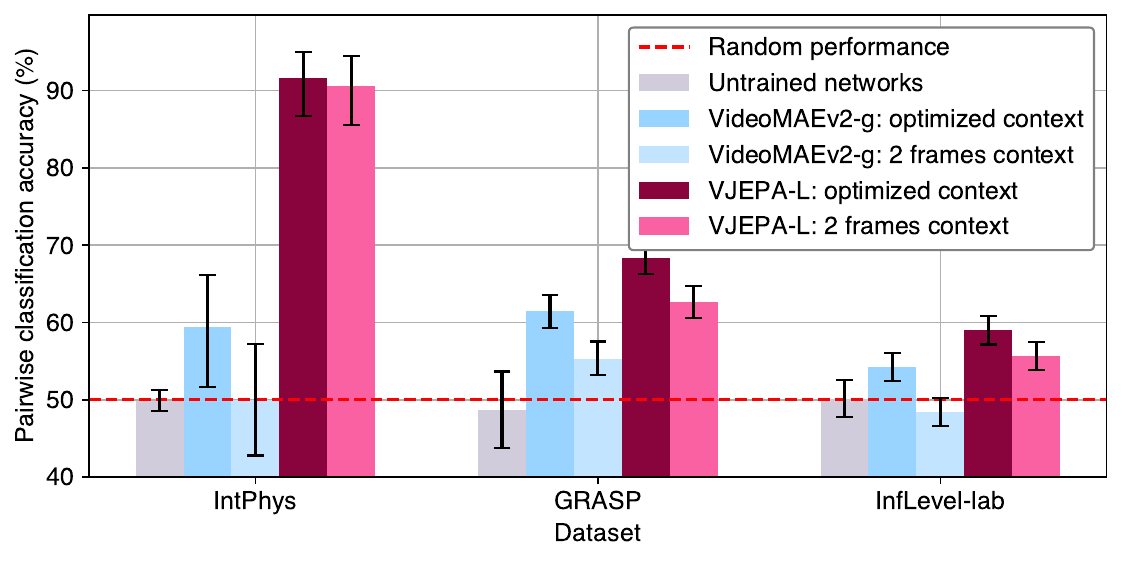}
    \caption{\textbf{Models perform suboptimally with a fixed context size.} Due to limitations in how long of a video models can process, we find drops in performance when using a single context size across all properties and datasets. Performance remains non-trivial for V-JEPA in this scenario.}
    \label{fig:ablation_context}
\end{figure}

\begin{figure}
    \centering
    \includegraphics[width=\linewidth]{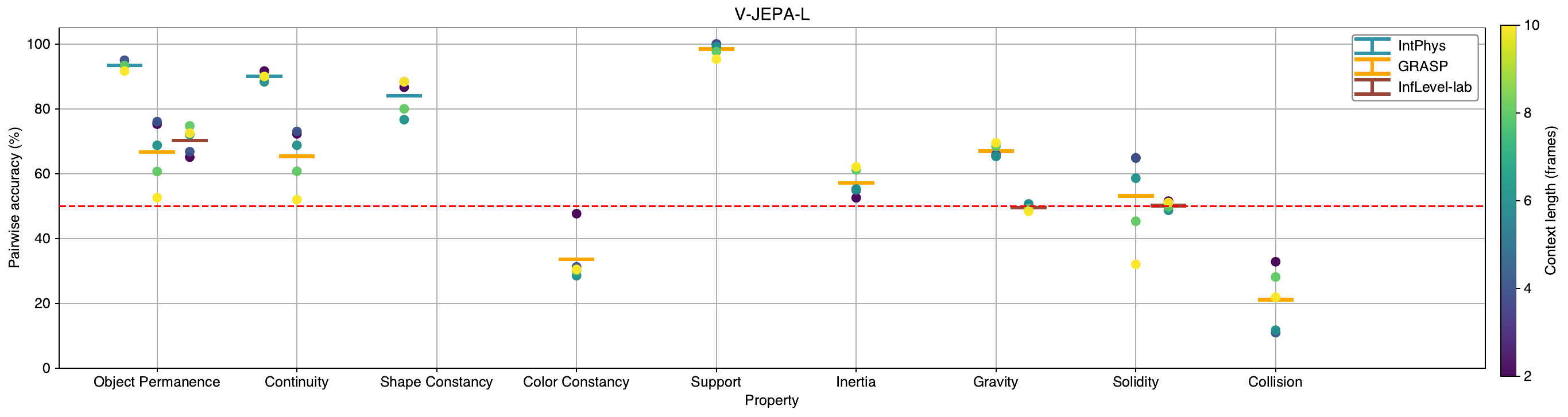}
     \includegraphics[width=\linewidth]{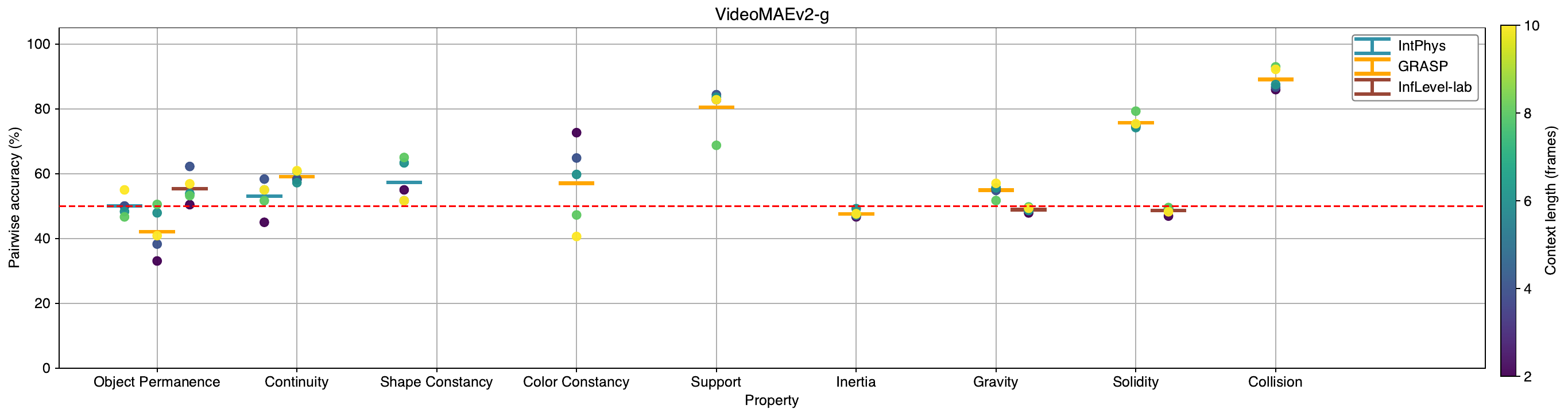}
    \caption{\textbf{Variation of performance when changing context size for predictions.} While models tend to perform better with smaller context sizes, we find the optimal context size to be dependent on the property and dataset. GRASP exhibits the most variation whereas IntPhys and InfLevel-lab are less sensitive in general.}
    \label{fig:ablation_context_methods}
\end{figure}

The fact that most datasets such as GRASP or InfLevel-lab do not come with ready-made validation and test split further complicates classical approaches. Even in their presence, the similarities of the scenes among a given property mean that a validation set would be too similar to the test set and cause information leakage. It is worth noting that the results obtained on the dev set of IntPhys correlate with the results on its private test set, as shown in Figures~\ref{fig:figure-1} and~\ref{fig:all_results}.

We thus opted to showcase the maximal attainable performance by optimizing the context size per property, as is commonly done in the self-supervised learning literature~\citep{bardes_vjepa_2024}. This means that we are evaluating whether the model has the necessary information or capabilities to solve the task at hand.

We now study this choice in more depth and its impact on performance. First, we study what happens with a fixed context size across all properties. Just as infants are not told which property is being tested in classical experiments~\citep{baillargeon_permanence_1991}, can the studied models perform well in this setting?
Second, we study the distribution of performance when the context size is varied for each property.

As illustrated in Figure~\ref{fig:ablation_context}, a fixed context size across properties and dataset can be used, with a minor impact on performance for V-JEPA. The small context size of 2 frames allows the model to do a longer term prediction of 14 frames. We hypothesize that this leads to better performance than the opposite, i.e. a 14 frame context, as it may be easier to predict some properties of the scene over a long horizon rather than remembering them from the past. To illustrate, if at a current frame a red ball is in frame, the long-term prediction may still include information about it. However, if it was shown at the beginning of the context and then hidden, the model may struggle to predict that the red ball is visible again. Further experiments would be necessary to better understand how models leverage their context.

Looking at Figure~\ref{fig:ablation_context_methods}, we obtain a better understanding of how context influences performance for specific properties and dataset.  Performance on IntPhys and InfLevel-lab is stable for different contexts for both V-JEPA and VideoMAEv2.
However, GRASP leads to the largest variations, which is also observed with untrained networks. We find that for the majority of properties, having a shorter context, and thus longer predictions, improves performance. This may be explained by biases in the design of GRASP, as well as biases in the models' predictions.

The variation in performance using different context sizes is a limitation of existing methods, which future models should address by handling longer video sequences.

\section{Influence of semantic and motion diversity in videos}

While we investigated the impact of the size of the pretraining dataset on performance, there are multiple ways to do so. Sampling only a part of the videos forming a dataset is one way, but we can also keep all videos and subsample the frames inside the each.

One way to see the difference between the two approaches is that, at a fixed total size, the former reduces the diversity of videos in terms of scenes, while the latter reduces the diversity in motion for a given scene. Thus, there is no reason why both should be equivalent.

To investigate frame subsampling, we use the following protocol: take the middle X\% of a video. Here, choosing the middle frames is not innocuous. The beginning and end of videos, especially tutorial videos~\citep{miech2019howto100m}, often consist of an intro, respectively an outro. These sections are not as related to the content of the videos and contain fewer movements and actions. We thus choose the middle frames as a way to get more meaningful data.

For video subsampling, the protocol is even simpler: sample uniformly X\% of the videos. In both cases, we ensure that every smaller set of videos in included in the bigger ones. If $X < Y$ the set of X\% of videos/frames is included in the set of Y\% of videos/frames.

\begin{figure}
    \centering
    \includegraphics[width=1\linewidth]{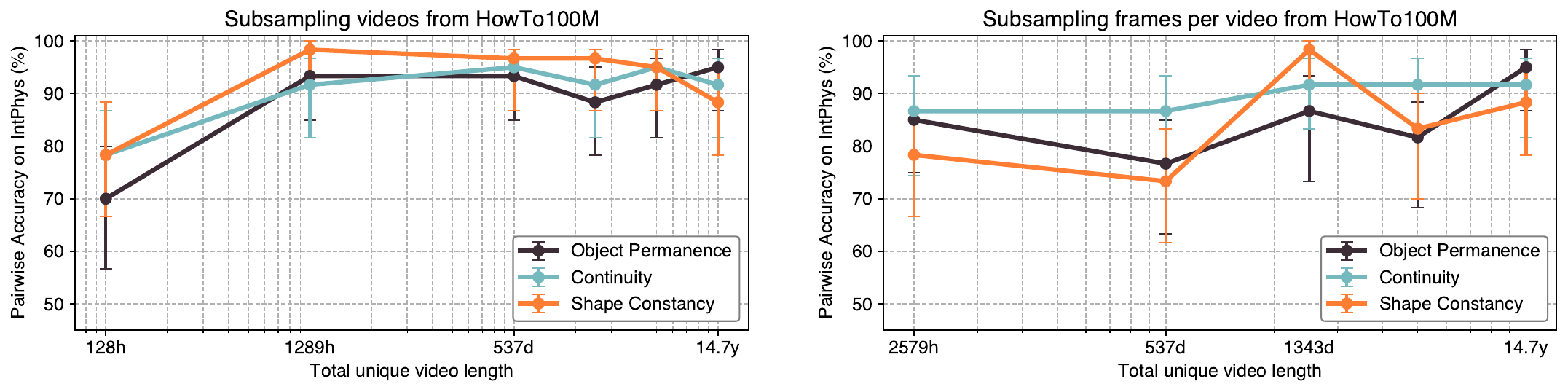}
    \caption{\textbf{Influence of motion and scene diversity.} By pretraining V-JEPA-L on subsets of HowTo100M, we investigate how the diversity in motion and scenes affects performance on IntPhys.\textbf{(left)} By subsampling videos, we reduce the diversity in scenes, where we find that the model can still reach good performance with 128h of unique videos. \textbf{(right)} By subsampling frames in videos, we reduce the diversity of motions in each scene. Here we find lower performance than when subsampling videos, but the model still achieves good performance with 2\% of the frames (2579h). }
    \label{fig:ablations-subsampling}
\end{figure}

In Figure~\ref{fig:ablations-subsampling} we can see that while both subsampling strategies lead to non-trivial performance, subsampling videos tend to produce higher performance. This further reinforces that all pretraining distributions are not equal and that certain ones lend themselves better to learn an understanding of intuitive physics.

\section{Results on the IntPhys challenge}

\begin{table}
\caption{\textbf{Pairwise error rates on IntPhys' test set.} For pairs of videos, taking either the maximum or average surprise from a video leads to high performance, surpassing the human results reported in~\citep{riochet_intphys_2022}.}
\label{tab:test-set-rel}
\resizebox{\textwidth}{!}{%
\begin{tabular}{llccccccccc}
\hline
Method & Surprise & \multicolumn{3}{c}{Object permanence}& \multicolumn{3}{c}{Shape constancy}&\multicolumn{3}{c}{Continuity}\\
       &                              & Visible & Occluded & All        & Visible & Occluded & All      & Visible & Occluded  & All    \\
\hline
V-JEPA-H & Max        & 0.6\% & 8.2\%   & 4.4\%      & 0.8\%  & 8.1\%   & 4.4\%   & 0.19\%  & 25.6\%   & 12.87\% \\
V-JEPA-H  & Avg                           & 0.0\%    & 0.56\%   & 0.28\%     & 0.0\%   & 0.0\%    & 0.0\%    & 0.0\%   & 0.19\%    & 0.09\% \\
V-JEPA-L & Max      & 5.2\%  & 35.4\%   & 0.20\%  & 8.8\%   & 35.0\%  & 21.9\% & 5.9\%   & 41.5\%  & 23.8\% \\
V-JEPA-L      & Avg                         & 0.9\% & 1.8\%   & 1.4\%  & 2.5\%  & 3.5\%  & 3.1\% & 0.7\% & 3.3\%  & 2.0\% \\
\cite{riochet2020occlusion}  &         & 5.0\%   & 19.0\%     & 12.0\%     & 11.0\%    & 31.0\%     & 21.0\%   & 26.0\%    & 47.0\%      & 41.0\%  \\
Human  &                               & 10.0\%   & 15.0\%     & 12.5\%     & 13.0\%    & 16.0\%     & 14.5\%   & 20.0\%    & 40.0\%      & 30.0\%  \\
\hline
\end{tabular}
}

\end{table}

\begin{table}
\caption{\textbf{Single video classification error rates (1-AUROC) on IntPhys' test set.} For single videos, we see that the maximum surprise of video leads to the highest performance, surpassing the human baselines reported in~\citep{riochet_intphys_2022}. Here, the average surprise of a video is not a good metric, possibly due to values being too dependent on the experimental setup. We report the metric as percentages for legibility.}
\label{tab:test-set-abs}
    \resizebox{\textwidth}{!}{%
\begin{tabular}{llccccccccc}
\hline
Method & Surprise & \multicolumn{3}{c}{Object permanence}& \multicolumn{3}{c}{Shape constancy}& \multicolumn{3}{c}{Continuity}        \\
       &                             & Visible & Occluded & All       & Visible  & Occluded & All     & Visible & Occluded & All    \\
\hline
V-JEPA-H & Max      & 8.0\%  & 28.1\%   & 19.2\%  & 11.9\%   & 29.7\%  & 21.9\% & 7.8\%   & 43.9\%  & 29.67\% \\
V-JEPA-H       & Avg                         & 27.8\% & 38.9\%   & 38.3\%  & 31.2\%  & 39.3\%  & 39.2\% & 28.4\% & 31.3\%  & 37.05\% \\
V-JEPA-L & Max      & 25.5\%  & 47.8\%   & 40.0\%  & 29.9\%   & 47.8\%  & 41.8\% & 26.0\%   & 49.0\%  & 41.6\% \\
V-JEPA-L       & Avg                         & 33.4\% & 41.7\%   & 41.5\%  & 37.0\%  & 42.5\%  & 42.7\% & 34.4\% & 38.8\%  & 41.5\% \\
Human  &                          & 18.0\%    & 30.0\%      & 24.0\%     & 22.0\%     & 30.0\%     & 26.0\%    & 28.0\%    & 47.0\%     & 38.0\%  \\
\hline
\end{tabular}
}

\end{table}
IntPhys was originally introduced as a challenge, with a private test set. While this makes analysis harder to provide for every experiment on this test set, we provide an analysis of V-JEPA on it.
As can be seen in Tables~\ref{tab:test-set-rel} and~\ref{tab:test-set-abs}, the high accuracy obtained by V-JEPA on the development set of IntPhys is also present when looking at the private test set. It consists of 3600 videos per property, compared to the 120 of the development set. These results reinforce the conclusions previously drawn, confirming the robustness of the learned intuitive physics understanding.\\
Similar to what was visible in Figure~\ref{fig:all_results}.B, V-JEPA is able to achieve performance similar or higher when compared to human baselines. We further find that V-JEPA surpasses the performance of previously published methods~\citep{riochet2020occlusion} which leverage pre-defined abstractions such as segmentation masks.

Focusing on the single video classification setting, we find a notable difference of performance between V-JEPA-H and V-JEPA-L. Where V-JEPA-H matches human peformance, V-JEPA-L remains far from it. This highlights the fact that scale can be beneficial when the task is harder, whereas it does not bring significant improvements in a pairwise setting.

\section{Importance of contextualization events on InfLevel}

As discussed in the main text, we find that models struggle on InfLevel-lab due to the importance of a contextualization event. Without knowledge and memory of it, the task becomes unsolvable as the occluder's modifications are invisible during the main experiment. We thus propose to relabel the data, assuming that objects are unmodified: a cup always has a bottom, and a cylinder always has an uncut back. This provides a testing ground that evaluates the understanding of the world of models where objects would be unmodified. 

There are some limitations, however. For gravity, for example, objects are seen going through a cup or not. Assuming that the cup always has a bottom, for the possible video nothing happens once the object is dropped, but for the impossible video, the model also needs to predict the trajectory of the object, bouncing on a table. This means that the impossible video is by default much harder than the possible one, which can skew performance upwards.

\begin{figure}
    \centering
    \includegraphics[width=0.8\linewidth]{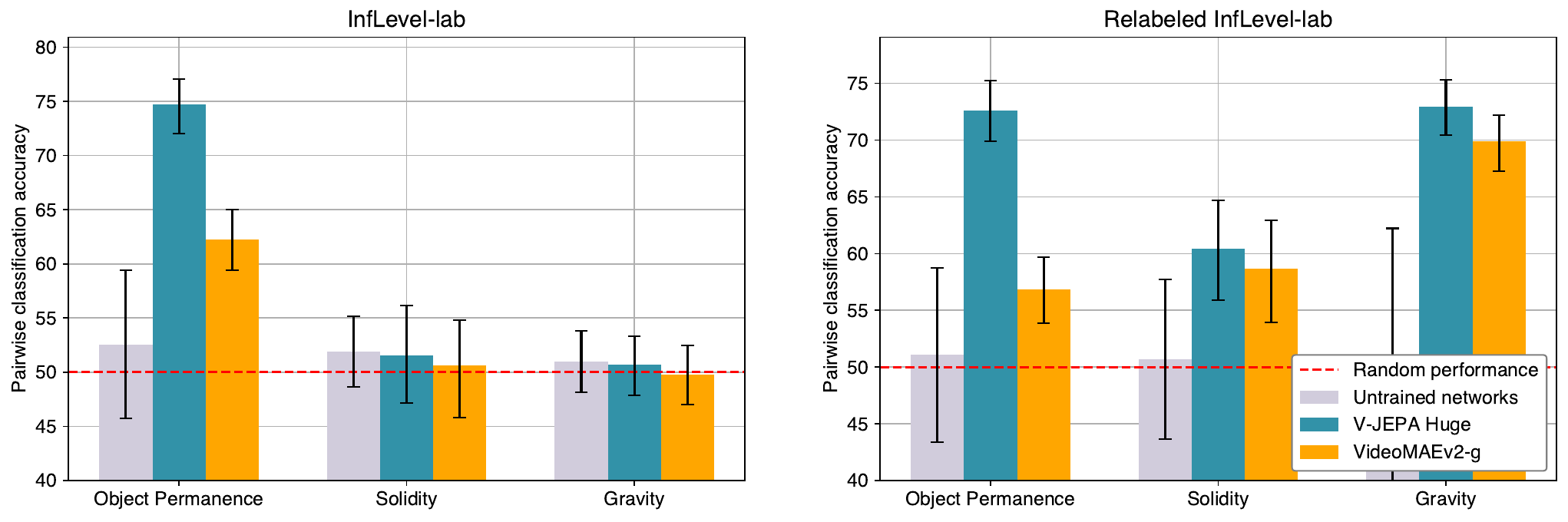}
    \caption{\textbf{Relabeling InfLevel to remove contextualization events.} Gravity and solidity both require to remember the properties about the containers shown in a video before the actual experiment. By relabeling the videos such that the prefix video is not necessary, we find a significant increase in performance for both V-JEPA and VideoMAE. However, this relabeling breaks the assumption that the possible and impossible videos have the same difficulty. }
    \label{fig:relabeled_inflevel}
\end{figure}

In Figure~\ref{fig:relabeled_inflevel}, we find a significant increase in performance for both V-JEPA and VideoMAEv2 but also for untrained networks. The increase for the latter models would suggest that the task can be solved with some heuristics, where the assumption that both videos are matched apart from a physics breaking event breaks.

As such, while this increase in performance is encouraging, it has to be taken with a grain of salt, and the results on continuity remain the most important due to the more controlled nature of the setup.

\section{Per-property performance of methods\label{sec:complete_all_methods}}

\begin{figure}[!h]
    \centering
    \includegraphics[width=1\linewidth]{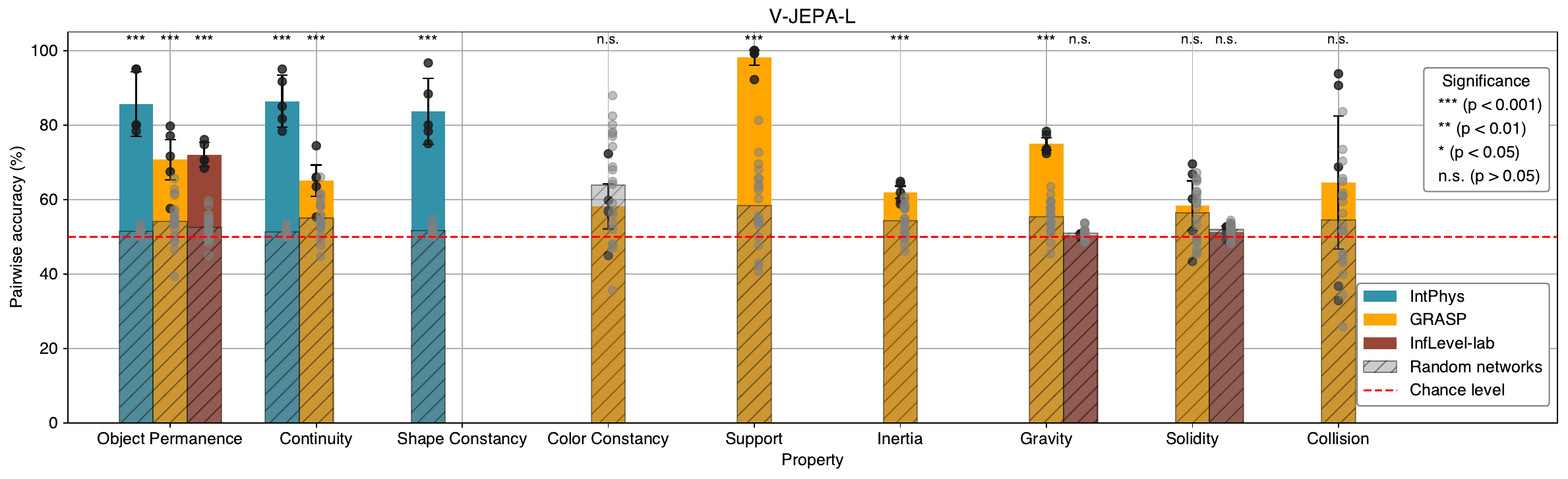}
    \caption{\textbf{Complete results for V-JEPA-L.} The models ($n=5$) achieve accuracies higher than untrained networks on most properties. Black dots represent the performance of 5 seeds.}
    \label{fig:complete-vjepa-l}
\end{figure}

\begin{figure}[!h]
    \centering
    \includegraphics[width=1\linewidth]{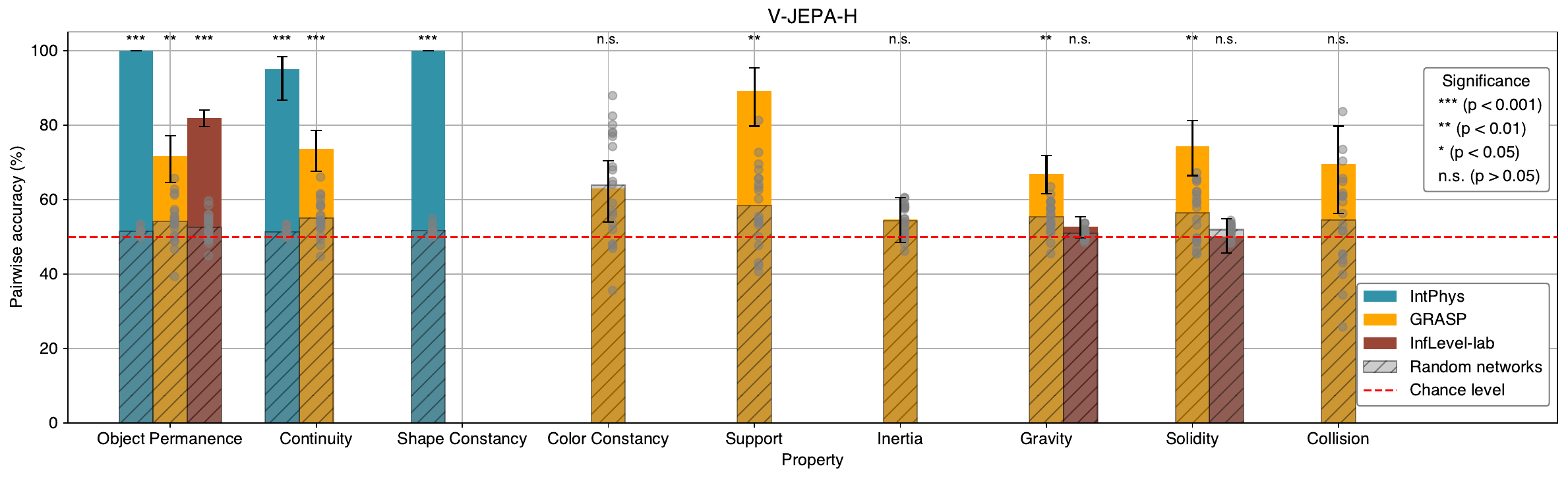}
    \caption{\textbf{Complete results for V-JEPA-H.} The model achieves accuracies higher than untrained networks on most properties. Gray dots represent the performance of the 20 untrained networks. Confidence intervals obtained via bootstrapping.}
    \label{fig:complete-vjepa-h}
\end{figure}

\begin{figure}
    \centering
    \includegraphics[width=1\linewidth]{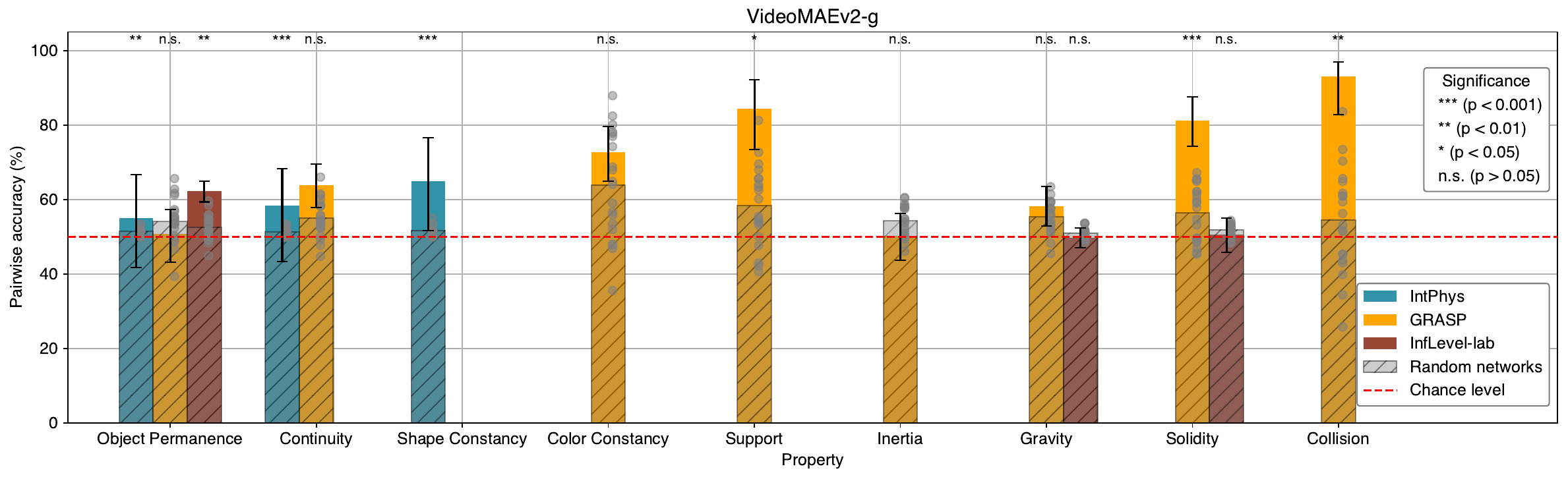}
    \caption{\textbf{Complete results for VideoMAEv2.} The model achieves performance on par or slightly higher than untrained networks across properties, apart from solidity and collision. Gray dots represent the performance of the 20 untrained networks. Confidence intervals obtained via bootstrapping.}
    \label{fig:complete-videomae}
\end{figure}

\begin{figure}
    \centering
    \includegraphics[width=1\linewidth]{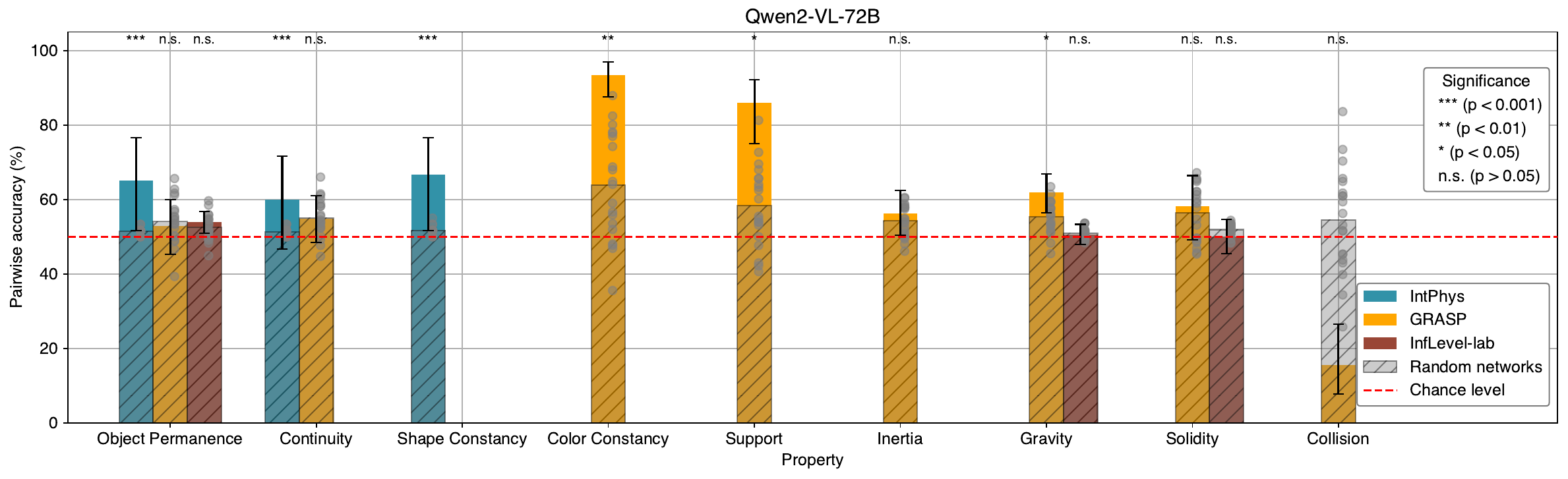}
    \caption{\textbf{Complete results for Qwen2-VL-72B.} The model achieves performance on par or slightly higher than untrained networks across properties, except for color constancy and support. Gray dots represent the performance of the 20 untrained networks. Confidence intervals obtained via bootstrapping.}
    \label{fig:complete-qwen}
\end{figure}

\begin{figure}
    \centering
    \includegraphics[width=1\linewidth]{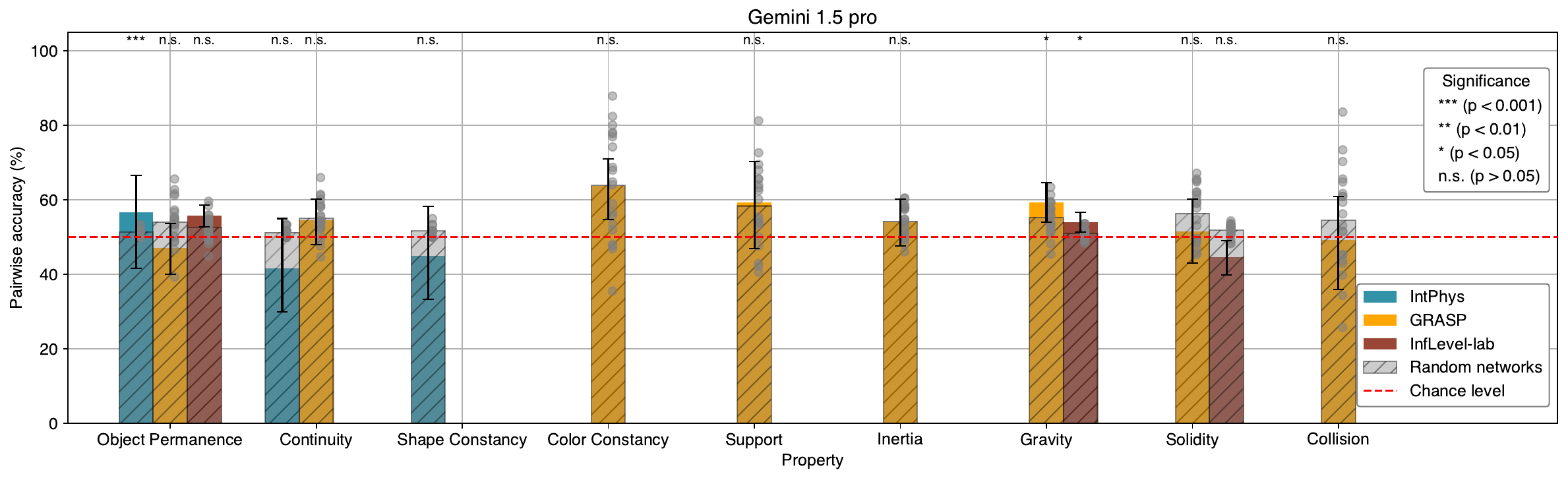}
    \caption{\textbf{Complete results for Gemini 1.5 pro.} The model achieves performance on par or slightly higher than untrained networks across properties. Gray dots represent the performance of the 20 untrained networks. Confidence intervals obtained via bootstrapping.}
    \label{fig:complete-gemini}
\end{figure}

\end{document}